\documentclass{article}

\usepackage{microtype}
\usepackage{graphicx}
\usepackage{subfigure}
\usepackage{booktabs} 
\usepackage{amsmath}
\usepackage{amssymb}
\usepackage{caption}
\usepackage{paralist}
\usepackage{scalefnt}
\usepackage{xspace}
\usepackage{natbib}
\usepackage{hyperref}

\usepackage[accepted]{icml2018}

\icmltitlerunning{Decoupling exploration and exploitation in deep RL}

\newcommand{\ddpg}{\textsc{ddpg}\xspace}

\newcommand{\acer}{\textsc{acer}\xspace}
\newcommand{\gep}{\textsc{gep}\xspace}
\newcommand{\geps}{\textsc{gep}s\xspace}
\newcommand{\imgeps}{\textsc{imgep}s\xspace}

\newcommand{\geppg}{\textsc{gep-pg}\xspace}
\newcommand{\qprop}{\textsc{q-prop}\xspace}
\newcommand{\ppo}{\textsc{ppo}\xspace}
\newcommand{\trpo}{\textsc{trpo}\xspace}
\newcommand{\sac}{\textsc{sac}\xspace}
\newcommand{\acktr}{\textsc{acktr}\xspace}

\newcommand{\dqn}{\textsc{dqn}\xspace}

\newcommand{\gps}{\textsc{gps}\xspace}
\newcommand{\hc}{HC\xspace}
\newcommand{\cmc}{Continuous Mountain Car\xspace}
\newcommand{\relu}{\textsc{relu}\xspace}
\newcommand{\adam}{\textsc{adam}\xspace}
\newcommand{\ql}{{\sc q-learning}\xspace}
\newcommand{\sota}{state-of-the-art\xspace}
\newcommand{\ks}{\textsc{ks}\xspace}
\newcommand{\ttest}{t-test\xspace}
\newcommand{\bs}{\textsc{bs}\xspace}

\definecolor{myred}{rgb}{0.8,0,0}
\definecolor{mygreen}{rgb}{0,0.6,0}
\definecolor{myblue}{rgb}{0,0,0.7}

\begin{document}


\twocolumn[
\vspace{-0.7cm}
\icmltitle{{GEP-PG}: Decoupling Exploration and Exploitation\\ in Deep Reinforcement Learning Algorithms}


\begin{icmlauthorlist}
\icmlauthor{C\'edric Colas}{in}
\icmlauthor{Olivier Sigaud}{in,up}
\icmlauthor{Pierre-Yves Oudeyer}{in}
\end{icmlauthorlist}

\icmlaffiliation{in}{INRIA, Flowers Team, Bordeaux, France}
\icmlaffiliation{up}{Sorbonne University, ISIR, Paris, France}

\icmlcorrespondingauthor{C\'edric Colas}{cedric.colas@inria.fr}

\icmlkeywords{deep reinforcement learning, goal exploration processes, exploration, continuous actions, sample efficiency}

\vskip 0.3in
]

\newcommand{\fix}{\marginpar{FIX}}
\newcommand{\new}{\marginpar{NEW}}


\printAffiliationsAndNotice{}  

\begin{abstract}
  In continuous action domains, standard deep reinforcement learning algorithms like \ddpg suffer from inefficient exploration when facing sparse or deceptive reward problems.
  Conversely, evolutionary and developmental methods focusing on exploration like Novelty Search, Quality-Diversity or Goal Exploration Processes explore more robustly but are less efficient at fine-tuning policies using gradient-descent.
  In this paper, we present the \geppg approach, taking the best of both worlds by sequentially combining a Goal Exploration Process and two variants of \ddpg. We study the learning performance of these components and their combination on a low dimensional deceptive reward problem and on the larger Half-Cheetah benchmark. We show that \ddpg fails on the former and that \geppg improves over the best \ddpg variant in both environments. Supplementary videos and discussion can be found at \url{frama.link/gep_pg}, the code at \url{github.com/flowersteam/geppg}.
\end{abstract}

\section{Introduction}

Deep reinforcement learning (RL) algorithms combine function approximation based on deep neural networks with RL. Outstanding results have been obtained recently by such systems in problems with either large discrete state and action sets \citep{mnih2015human,silver2016mastering} or continuous state and action domains \citep{lillicrap2015continuous}. These results have attracted considerable attention on these techniques in the last two years, with the emergence of several competitive algorithms like \ppo \citep{schulman2017proximal}, \acktr \citep{wu2017scalable}, \acer \citep{wang2016sample} and \qprop \citep{gu2016q}, as well as easily accessible benchmarks \citep{brockman2016openai,machado2017revisiting} to perform thorough comparative evaluations \citep{duan2016benchmarking,islam2017reproducibility,henderson2017deep}.

Deep RL algorithms generally consist in applying Stochastic Gradient Descent (SGD) to the weights of neural networks representing the policy and an eventual critic, so as to minimize some reward-related cost function. 
The good performance of these methods mainly comes from the high sample efficiency of SGD, the gradient with respect to the network parameters being analytically available. 
To deal with the exploration-exploitation trade-off \citep{sutton98}, a deep RL algorithm uses its current policy to select the best action (exploitation), and adds random noise to explore (action perturbation). This could be a simple Gaussian noise or a more advanced Ornstein-Uhlenbeck (OU) correlated noise process in the case of Deep Deterministic Policy Gradient \ddpg \citep{lillicrap2015continuous} or a random selection of suboptimal actions in the case of \dqn \citep{mnih2015human}. This approach may not be efficient enough in the case of RL problems with multi-dimensional continuous actions where exploration can be particularly challenging (e.g. in robotics).


Recently, a family of evolutionary computation techniques has emerged as a convincing competitor of deep RL in the continuous action domain \citep{salimans2017evolution,conti2017improving,such2017deep}.
These techniques perform policy search directly in the policy parameter space, classifying them as ``parameter perturbation'' methods.
These methods benefit from more focus on exploration, especially for problems with rare or deceptive rewards and continuous action spaces, as illustrated by many papers on Novelty Search \citep{lehman2011abandoning,conti2017improving}, Quality-Diversity \citep{pugh2015confronting,cully2017quality} and curiosity-driven learning \citep{baranes_active_2013, forestier2017intrinsically}.
Besides, policy parameter perturbation is generally considered superior to action perturbation in policy search \citep{stulp13paladyn}.
But, as they do not rely on analytical gradient computations, these methods are generally less sample efficient than SGD-based methods \citep{salimans2017evolution}.

To summarize, deep RL methods are generally more efficient than evolutionary methods once they are fine-tuning a solution that is already close to the optimum, but their exploration might prove inefficient in problems with flat or deceptive gradients.

\newpage
\textbf{Contributions.} In this paper, we define a new framework which takes the best of both worlds by sequentially combining elements of both families, focusing on its use and evaluation for continuous action RL problems.
We call it \geppg for ``Goal Exploration Process - Policy Gradient''.
In the first stage of \geppg, a method coming from the curiosity-driven learning literature called Goal Exploration Processes (\geps) \citep{forestier2017intrinsically} is used to efficiently explore the continuous state-action space of a given problem. The resulting samples are then stored into the replay buffer of a deep RL algorithm, which processes them to perform sample efficient policy improvement. In this paper, the chosen deep RL algorithm is \ddpg, which can itself benefit from various exploration strategies \cite{lillicrap2015continuous,plappert2017parameter}.

For evaluation purposes, we use two benchmarks showing distinct properties: 1) the \cmc (CMC) environment, a small size ($2D$ state, $1D$ action) control benchmark with interesting deceptive reward properties and 2) Half-Cheetah (\hc), a larger ($17D$ state, $6D$ action) benchmark where \ddpg was recently the \sota (now behind \textsc{sac}\xspace \citep{haarnoja2018soft}).

On these benchmarks, we study the learning performance of \gep, variants of \ddpg, and of the sequential combination of both components.
We study the learning performances from several point of views: final performance, sample efficiency to reach high-performance, and variability of the learning dynamics.
These investigations reveal various aspects of the exploration and exploitation efficiency of these components.
Overall, we finally show that \geppg provides results beyond \ddpg in terms of performance, variability and sample efficiency on \hc. 

The paper is organized as follows. In Section~\ref{sec:related}, we identify three categories of methods to perform exploration in policy search and mention research works sharing some similarities with ours. In Section~\ref{sec:methods}, we quickly describe \ddpg, then \geps, the way we combine them into \geppg, and the experimental setup and parameters. Performance, variance and sample efficiency results of the corresponding exploration strategies on CMC and \hc are given in Section~\ref{sec:results}. Finally, we discuss these results, conclude and describe potential avenues for future work in Section~\ref{sec:conclusion}.

\section{Related work}
\label{sec:related}
As deep RL algorithms start addressing larger and more difficult environments, the importance of exploration, particularly in the case of sparse reward problems and continuous actions, 
is a more and more recognized concern.

Basic, {\em undirected exploration} methods rely mostly on noise.
This is the case in methods like \trpo \citep{schulman2015trust}, \ppo \citep{schulman2017proximal} and \sac \citep{haarnoja2018soft} which use a stochastic policy, but also in methods using a deterministic policy like \ddpg, which add noise to the chosen action.
Another approach consists in adding noise directly into the neural networks used as representation for learning \citep{fortunato2017noisy,plappert2017parameter}. The latter approach is generally considered superior \citep{stulp13paladyn}, even if this can depend on the respective sizes of the policy parameter space and the state action space.

A more directed exploration strategy strives to cover the state-action space as uniformly as possible. This idea has been studied both in the deep RL and in the evolutionary/developmental literatures. 
In deep RL, this is achieved using count-based exploration \citep{bellemare2016unifying}, as well as various forms of intrinsic rewards and exploration bonuses \citep{houthooft2016curiosity,houthooft2016vime,pathak2017curiosity} and methods for option discovery \citep{machado2017laplacian}. Count-based exploration can also be applied in a more compact {\em latent space} obtained from representation learning techniques \citep{tang2016exploration}. While some of these strategies have been tested for continuous actions RL problems \citep{houthooft2016vime, tang2016exploration}, they remain complex to implement and computationally expensive. Besides, exploration bonuses focus on information gain associated to states (rather than state-action), thus they do not fully consider the exploration challenge of multi-dimensional continuous action spaces. 

Evolutionary and developmental methods focused on exploration, like Novelty Search \citep{lehman2011abandoning,conti2017improving}, Quality-Diversity \citep{pugh2015confronting,cully2017quality} and Curiosity-Driven Learning \citep{baranes_active_2013,forestier2016curiosity} generalize the idea of covering well the state-action space. Indeed, instead of trying to cover the state-action space of a predefined RL problem, they try to cover a user-defined space of behavioral features, called ``outcome space" or ``behavior space", potentially characterizing the full state-action trajectories resulting from a policy roll-out. Within approaches of curiosity-driven exploration, Goal Exploration Processes are methods where this outcome space is used to stochastically sample goals, enabling to efficiently discover repertoires of policies that produce diverse outcomes. Learning such diverse repertoires of policies is often useful for finding approximate solutions to difficult problems with flat, deceptive or rare rewards.
However, these methods are episode-based rather than step-based \citep{deisenroth2013survey}, 
hence they are generally less sample efficient in fine-tuning the parameters of a policy (see Section~\ref{sec:geps} for more details).


Our approach sequentially combines evolutionary/ developmental methods for exploration and more traditional deep RL algorithms for fine-tuning policy parameters. This is similar to \cite{nair2017overcoming}, in which demonstration trajectories are used to inform \ddpg through the replay buffer.
Using Goal Exploration Processes provides extra information on what might be useful to explore via the design of the outcome space, unlike imitation learning which provides more directive information on how to behave.
Besides, like the {\em Guided Policy Search} (\gps) algorithm \citep{levine2013guided}, our approach can first use simple policies to generate samples and then train a richer policy.
Finally, like us, \citet{zimmer2017bootstrapping} first explore an environment with an evolutionary method, then call upon an RL algorithm to speed up policy convergence. However, they do so by extracting a discrete state-action representation and using the \ql algorithm \citep{watkins89}, avoiding the burden of solving continuous action problems.

\section{Methods}
\label{sec:methods}
In this section we quickly define \ddpg and Goal Exploration Processes (\geps) as background methods of our study, then we present \geppg and we describe the experimental setup. Our code-base is made available online\footnote{https://github.com/flowersteam/geppg}.

\subsection{\ddpg}
\label{sec:ddpg}

The \ddpg algorithm \citep{lillicrap2015continuous} is a deep RL algorithm based on the Deterministic Policy Gradient \citep{silver2014deterministic}. It borrows the use of a replay buffer and a target network from \dqn \citep{mnih2015human}. In this paper, we use two versions of \ddpg: 1) the standard implementation of \ddpg with action perturbations ($OU$ noise) \citep{lillicrap2015continuous}; 2) a variant in which the neural network used as actor is augmented with some noise (parameter perturbations), resulting in better exploration and more robust performance \citep{plappert2017parameter}. All the parameters of \ddpg used in this paper are taken from baselines used in systematic studies \citep{islam2017reproducibility,henderson2017deep}, and are described in Section~\ref{sec:exp_params}. 

\subsection{Goal Exploration Processes}
\label{sec:geps}

Goal Exploration Processes (\geps) are a subpart of Intrinsically Motivated Goal Exploration Processes (\imgeps), a more general framework addressing efficient exploration and curriculum learning in developmental scenarios \citep{forestier2017intrinsically}. Here we focus on the \gep part, describing it under the perspective of a policy search process endowed with efficient exploration properties.

In \geps, two spaces are used: the policy parameter space $\Theta$ and the outcome space $O$. The latter is designed by the experimenter, and is often a set of behavioral features characterizing the state-action trajectory produced by running a policy in the environment. Thus, the outcome $o$ (which is a vector of features) of a policy parameterized by $\theta$ can be measured by the system after each roll-out of this policy. 

The general process can be described as follows.
In a first {\em bootstrap} stage, $N$ sets of policy parameters $\theta$ are drawn randomly from $\Theta$. For each policy the resulting outcome $o$ in $O$ is observed. This stage implements a pure random policy search process.
Both at this stage and at the next one, each time some new point $o$ is observed, the corresponding $\langle\theta,o\rangle$ pair is stored.
In the second stage, vectors $o$ are randomly drawn from $O$.
These points are considered as goals and are not related to the actual goal of a particular benchmark. \gep therefore implements an exploration that is {\em directed} towards the generation of behavioral diversity and is completely unaware of the external reward signal of a benchmark. 
Once these goals are drawn, the algorithm looks for policy parameters $\theta$ which could reach them using a simple search method. It looks for the closest stored point in the outcome space (a nearest-neighbor strategy) and randomly perturbs the corresponding $\theta$. If the perturbation is adequate by chance, the generated outcome should be closer to the goal than the previous one. Importantly, even if the perturbations is not closer to the goal, it enables to discover a novel outcome and progress towards other goals as a side effect. 
Instead of random perturbations, more advanced interpolation-based techniques can be used \citep{forestier2016curiosity}, but here we stick to the simplest method and apply Gaussian noise $\mathcal{N}(0\ ,\sigma^{2})$.

By storing more and more $\langle\theta,o\rangle$ pairs, \geps improve reaching capabilities to a variety of goals. As it stores a population of policies rather than trying to learn a single large goal-parameterized policy, and as generalization can be made on-the-fly through various forms of regression \citep{forestier2016curiosity}, this can be viewed as a form of {\em lazy learning} \citep{aha1997editorial}. Described this way, \geps share a lot of similarities with Novelty Search \citep{lehman2011abandoning,conti2017improving} and Quality-Diversity \citep{pugh2015confronting,cully2017quality} methods.

While \geps do not use a potential reward information corresponding to an externally imposed RL problem when they explore, e.g. in a benchmark, they can observe this RL-specific reward during policy roll-outs and store it for later reuse. In the \geppg framework described below, this is reused both by the learning algorithm to initialize a replay buffer for \ddpg and by the experimenter to measure the performance of the best policies found by \gep from the perspective of a target RL problem (and even if \gep does not try to optimize the associated reward).

\subsection{The \geppg methodology}
\label{sec:geepg}

The \geppg methodology comes with two steps. In a first {\em exploration} stage, a large diversity of samples are generated and stored using \gep, using either a simple linear policy or the same policies as \ddpg. In a second {\em exploitation} stage, \ddpg's replay buffer is loaded with \gep-generated samples and it is used to learn a policy. As described above, in this paper we use two versions of \ddpg coined {\em action-} and {\em parameter-} perturbation.

\subsection{Experimental setups}
\label{sec:setups}

Below we describe the two benchmark environments used, namely CMC and \hc.
For both setups, we use the implementations from OpenAI Gym \citep{brockman2016openai}.

\subsubsection{Continuous Mountain Car}
\label{sec:cmc}


The Mountain Car environment is a standard benchmark for the continuous state, discrete action RL setting. An underpowered car whose actions are in $\{-1,0,1\}$ must reach its goal at the top of a hill by gaining momentum from another hill. The state space contains the horizontal position and velocity and the action is a scalar acceleration.
In the standard setting, the goal is to reach the flag as fast as possible. A bang-bang policy, using only $\{-1,1\}$ as actions, is the optimal solution. The environment is reset after $10^3$ steps or when the flag is reached.

In the {\em \cmc} (CMC) environment, the action is defined over $[-1,1]$ and the car should reach the goal while spending as little energy as possible.
To ensure this, the reward function contains a reward of $100$ for reaching the goal and, at each step, an energy penalty for using acceleration $a$ which is $-0.1 \times a^2$. 
This reward function raises a specific exploration issue: it is necessary to perform large accelerations to reach the goal but larger accelerations also bring larger penalties. As a result, the agent should perform the smallest sufficient accelerations, which results in a ``Gradient Cliff'' landscape as described in \cite{lehman2017more}. If the goal is not reached, the least negative policy would consist in performing null accelerations. Thus, as long as the agent did not reach the goal, the environment provides a {\em deceptive} reward signal which may drive policy updates in the wrong direction. As a consequence, when using an SGD-based approach, successful learning is conditioned on reaching the goal fast enough,
so that the attractiveness of getting the goal reward overcomes the tendency to stop accelerating.
Due to all these features, CMC is a favorable environment for our study.

\subsubsection{Half-Cheetah}
\label{sec:hc}

Our second benchmark is \hc, in which the agent moves a 2D bipedal body made of 8 articulated rigid links. The agent can observe the positions and angles of its joints, its linear positions and velocities ($17D$) and can act on the torques of its legs joints ($6D$). The reward $r$ is the sum of the instantaneous linear velocities on the $x$ axis $v_x(t)$ at each step, minus the norm of the action vector $a$: \mbox{$r=\sum_t(v_x(t)-0.1\times|a(t)|^2)$}. The environment is reset after $10^3$ steps, without any other termination criterion.
As in CMC, \hc shows a ``Gradient Cliff'' landscape property as we experienced that a cheetah running too fast might fall, resulting in a sudden drop of performance close to the optimum. 

\subsection{Evaluation methodology}

Difficulties in reproducing reported results using standard algorithms on standard benchmarks is a serious issue in the field of deep RL. For instance, with the same hyper-parameters, two different codes can provide different results \citep{islam2017reproducibility,henderson2017deep}. In an attempt to increase reproducibility, the authors of \cite{henderson2017deep} have put forward some methodological guidelines and a standardized evaluation methodology which we tried to follow as closely as possible. To avoid the code issue, we used the implementation of Henderson's fork of OpenAI Baselines \citep{baselines} for all the variants of \ddpg.
In our work as in theirs, \ddpg is run in the environment for $100$ roll-out steps before being trained with $50$ random batches of samples from the replay buffer.
This cycle is repeated $20$ times ($2.10^3$ steps in the environment) before \ddpg is evaluated off-line on $10^4$ steps ($10$ episodes).

However, this standardized methodology still raises two problems. First, though the authors show that randomly averaging the results over two splits of 5 different random seeds can lead to statistically different results, they still perform evaluations with $5$ seeds. In all the experiments below, we use $20$ different seeds.
In order to make sure averaging over $20$ seeds was sufficient to counter the variance issue, we run $40$ trials of the baseline \ddpg with OU noise $(\mu=0,\sigma=0.3)$ for both environments and computed our comparison tests for $1000$ different sets of $20$ randomly selected runs.
To compare the performance of the various algorithms, we use the tests advised by \cite{henderson2017deep}: a paired t-test and a bootstrap estimation of the 95\% confidence interval of the means difference using $10^4$ bootstraps.
For all tests listed above, less than 5.01\% of the sets were found statistically distinct, with most of the test showing a 0\% error.
Two groups of $20$ seeds have therefore between 0\% and 5\% chances of being found separable depending on the considered test and metric (see Appendix A).
In all figures, we show the mean and standard error of the mean across $20$ runs using different random seeds while the x-axis is the number of time steps.

Second, the authors of \cite{henderson2017deep} report what we call a {\em final metric} of performance: the average over the last $100$ evaluation episodes, $10$ episodes for each of the last 10 controllers. This can be a problem for two reasons: 1) when progress is unstable, the final steps may not correspond to the overall best performing policies; 2) an average over several different policies obtained along learning does not mean much. Using this metric, only the last policies are considered, forgetting about the potential good ones discovered in the past, good policies that we might want to save for reuse.

As an alternative, we find it more appropriate and more informative to report the performance over $100$ evaluation episodes of the best policy found over the whole training process (called {\em absolute metric}). The process duration is set to $2M$ steps on \hc to keep close to the standard methodology, and $5.10^5$ steps on CMC. 

Concerning the \gep part of \geppg, we keep the same structure and show the average performance over 10 episodes ($10^4$ steps) every $2.10^3$ steps in the environment. Note that the termination condition of CMC implies that an episode does not always correspond to $10^3$ steps, e.g. when the goal is reached before the end. We therefore show the performance evaluated at the nearest step being a multiple of the figure resolution ($2.10^3$ steps). 

In the results below, we report the performance as evaluated by both the absolute and the final metrics, to facilitate comparison with the literature while providing more informative evaluations. 

\subsection{Experimental parameters}
\label{sec:exp_params}

For \ddpg, we use the default set of parameters provided in Henderson's version mentioned above. The replay buffer is a sliding window of size $10^6$. We also tried a replay buffer size of $2.10^6$ samples, but this led to lower performance in all cases. Three types of noise are used in our different comparisons: 1) an OU process with $\mu=0$, $\sigma=0.3$ and $\theta=0.15$, 2) an OU version where $\sigma$ linearly decreases from $0.6$ at the first step to $0$ at the final step, and 3) the noise on actor network parameters with variance $\sigma=0.2$, which shows \sota performance on \hc \citep{plappert2017parameter,henderson2017deep}.
The other meta-parameters of \ddpg are as follows: the batch size is $64$, the discount factor is $0.99$, the actor and critic networks have two hidden layers of size $(64,64)$ and \relu activation functions, the output layer activation function are $tanh$ and linear respectively. The learning rates are $10^{-4}$ and $10^{-3}$ respectively and the SGD algorithm is \adam.

As stated above, the simple \gep policies are linear policies with $tanh$ activation functions.
In CMC, it maps the position and velocity to the action and the corresponding search space is $2D$ instead of $4288D$ if we had used the \ddpg policy. In \hc, it maps the position and velocity of the joints to the torques (12 observations), it is $72D$ instead of $5248D$.
In a bootstrap stage, the \gep  policy parameters are sampled randomly in $[-1,1]^P$ where $P$ is the number of parameters.
Bootstrap consists of $5$ episodes for CMC, $50$ for \hc.
Then $45$ (respectively $450$) random goals are drawn in a custom goal space which is $3D$ for CMC (range of position, maximum position and energy spent) and $2D$ for \hc (mean velocity and minimum head position). Given a goal, the algorithm finds in its memory the closest state previously experienced using a k-nearest neighbor algorithm using an Euclidean distance and $(k=1)$, and samples parameters around the associated set of policy parameters with centered Gaussian noise using $\sigma=0.01$.

\section{Results}
\label{sec:results}

In this section, we proceed as follows. First, we investigate the exploration efficiency of action perturbation and policy parameter perturbation methods in \ddpg using the CMC and \hc benchmarks. Second, we examine whether \geps are better than the above undirected exploration methods at reaching the goal for the first time on CMC. Third, we use the \geppg algorithm to investigate whether the better exploration capability of \geps translates into a higher learning performance on CMC and \hc. We show that, though \gep alone performs better than \geppg on CMC, \geppg outperforms both \gep alone and the \ddpg variants on \hc, in terms of final performance and variance.

\subsection{Undirected exploration in \ddpg}
\label{sec:undirec}

\begin{figure*}[hbtp]
  \centering
 \subfigure[\label{noise_cmc}]{\includegraphics[width=0.65\columnwidth]{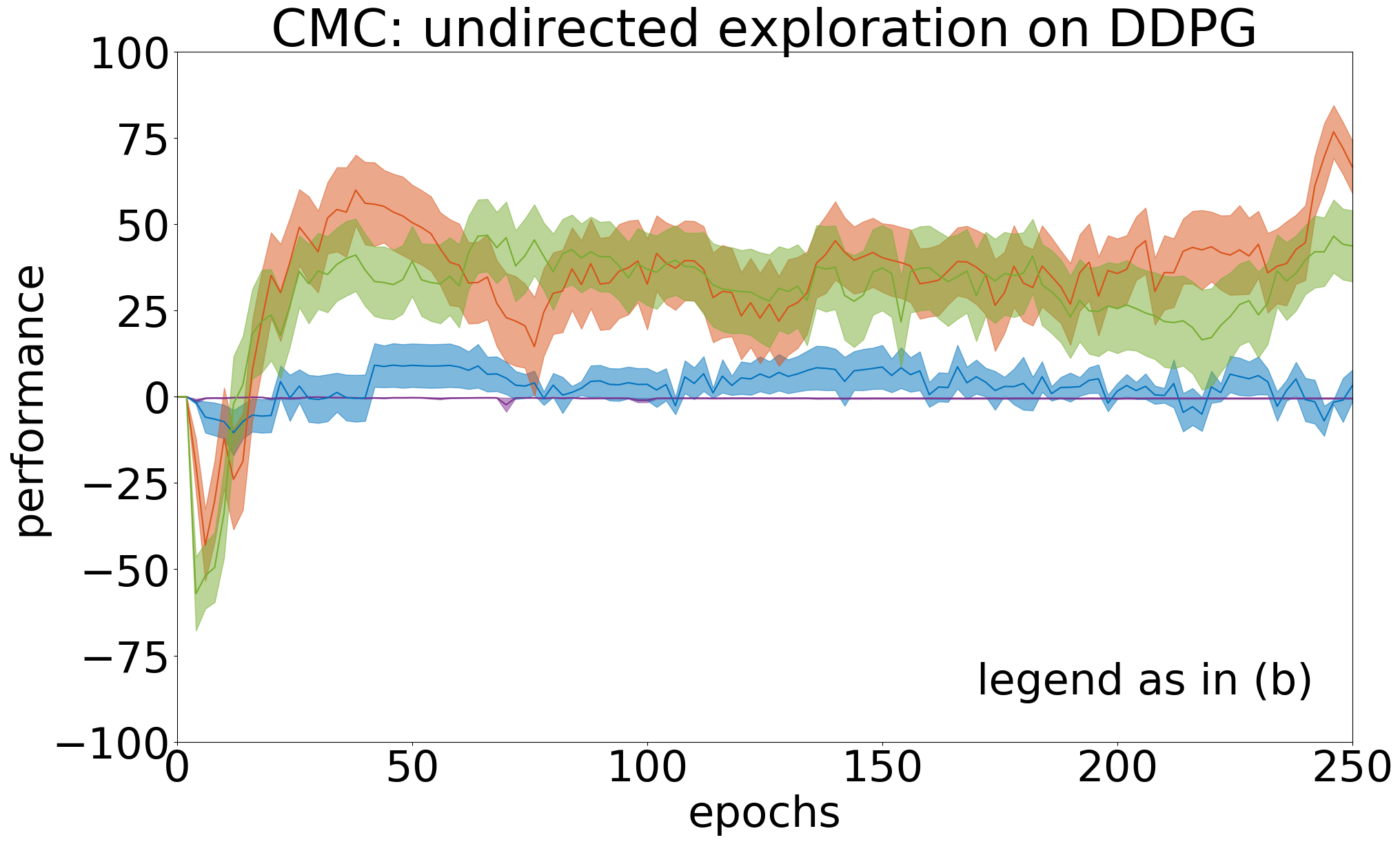}}
 \subfigure[\label{histo_noise_cmc}]{\includegraphics[width=0.63\columnwidth]{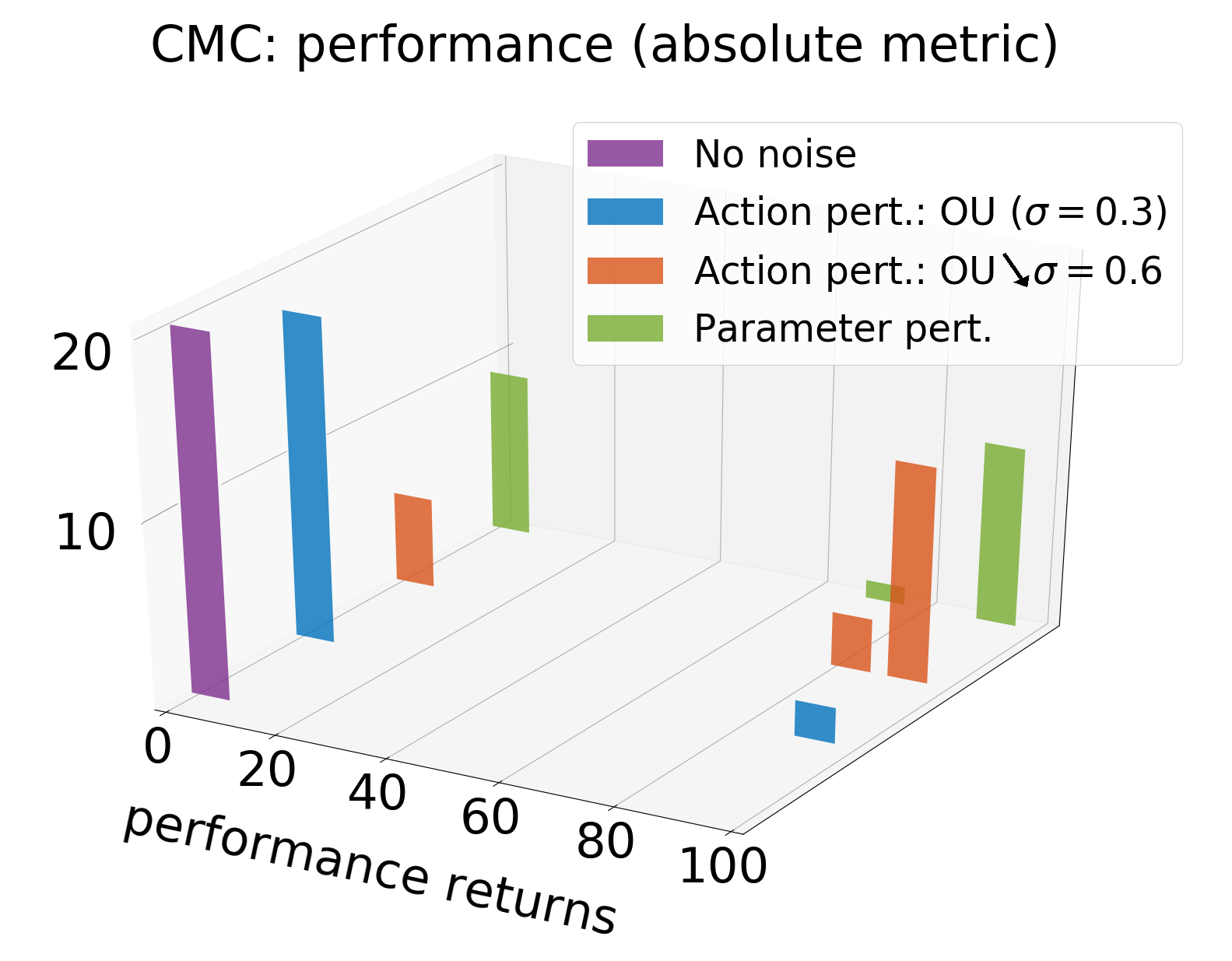}}
 \subfigure[\label{noise_hc}]{\includegraphics[width=0.65\columnwidth]{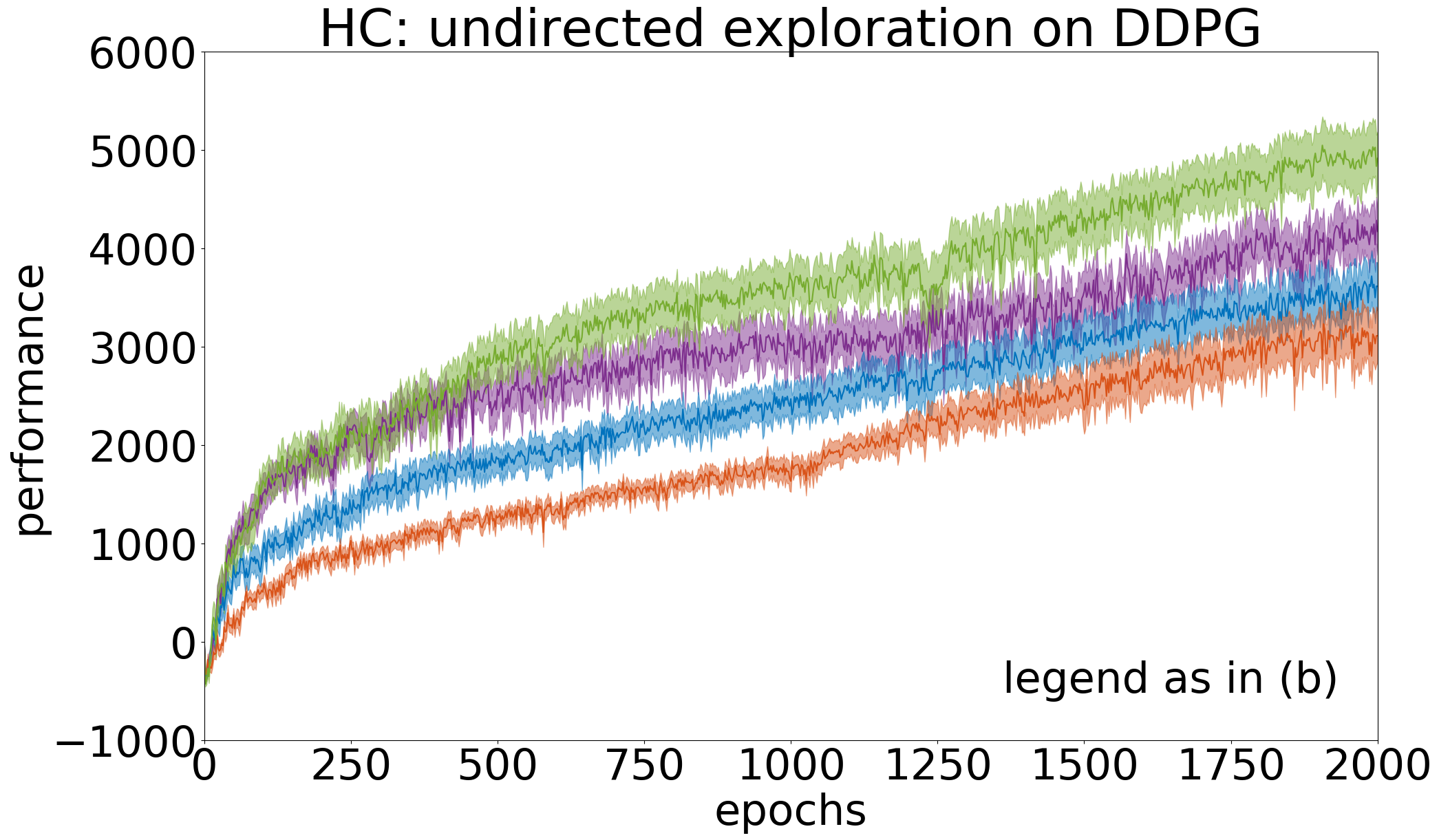}}
 \caption{(a): Effect of action perturbation versus policy parameter perturbation on the performance of \ddpg on CMC. (b): Histogram of absolute performances on CMC (c): Same as (a) for \hc. \label{fig:noise}}
\end{figure*}

First, we compare the effect of action perturbation versus policy parameter perturbation on \ddpg performance on CMC and \hc.
More precisely, we compare in \figurename~\ref{fig:noise}: 1) a \ddpg version without any exploration noise as a baseline, 2) the approach of \cite{lillicrap2015continuous} using OU noise with $(\mu=0, \sigma=0.3)$, 3) the parameter perturbation method implemented in \cite{plappert2017parameter} and 4) OU noise with $\sigma$ linearly decreasing from $0.6$ to $0$.

Consistently with \cite{plappert2017parameter}, we found that parameter perturbation outperforms action perturbation on \hc, but also on CMC on both metrics of performance (all tests significant at the 5\% level, see Appendix D).

On CMC, the \ddpg version without noise performs poorly as it does not have the exploration drive required to reach the reward a first time. The decreasing noise strategy significantly improves over the static one, as a stronger initial noise helps finding the goal more often.
On \hc, it seems that having no noise, a standard OU(0.3) or a decreasing noise do not lead to statistically different performances at 2M steps, although difference could be noted earlier on. We conclude that increasing noise in the first stage is beneficial in CMC but might be detrimental in \hc. 

These contrasting results might be due to several phenomena. First, there is probably less need to explore on \hc than on CMC, as reward information is available anytime.
Second, as explained in Section~\ref{sec:cmc}, reaching the goal early is crucial on CMC, which is not the case on \hc. Using a stronger exploration noise in the first stage might therefore bring a significant advantage on CMC. Third, it might be the case that too much noise is more detrimental to performance on \hc than on CMC because on \hc more noise may result in more falls.

However, the information depicted in \figurename~\ref{noise_cmc} must be taken with a grain of salt. Indeed, the curves depict averages over $20$ runs, but the individual performances are not normally distributed (see Appendix C). Because learning is unstable, an algorithm having found a good policy during learning might present a bad one in the final steps. To present complementary evaluations, we chose to report the histograms of absolute metrics across seeds. This absolute metric for a given seed is computed as the average performance over 100 test episodes of the best policy found across training, see \figurename~\ref{histo_noise_cmc}.

As a conclusion, parameter perturbations perform better than action perturbations, whereas increasing exploration in a first stage might be either advantageous or detrimental to the performance depending on the environment properties.

\subsection{Exploration efficiency of \gep and \ddpg on CMC}

As explained above, the CMC benchmark raises a specific exploration issue where the time at which the goal is first reached is crucial.
The same is not true of \hc.
In this section, we study how fast two variants of \gep and two variants of \ddpg can find the goal.

\begin{figure}[hbtp]
  \centering
    \includegraphics[width=0.80\columnwidth]{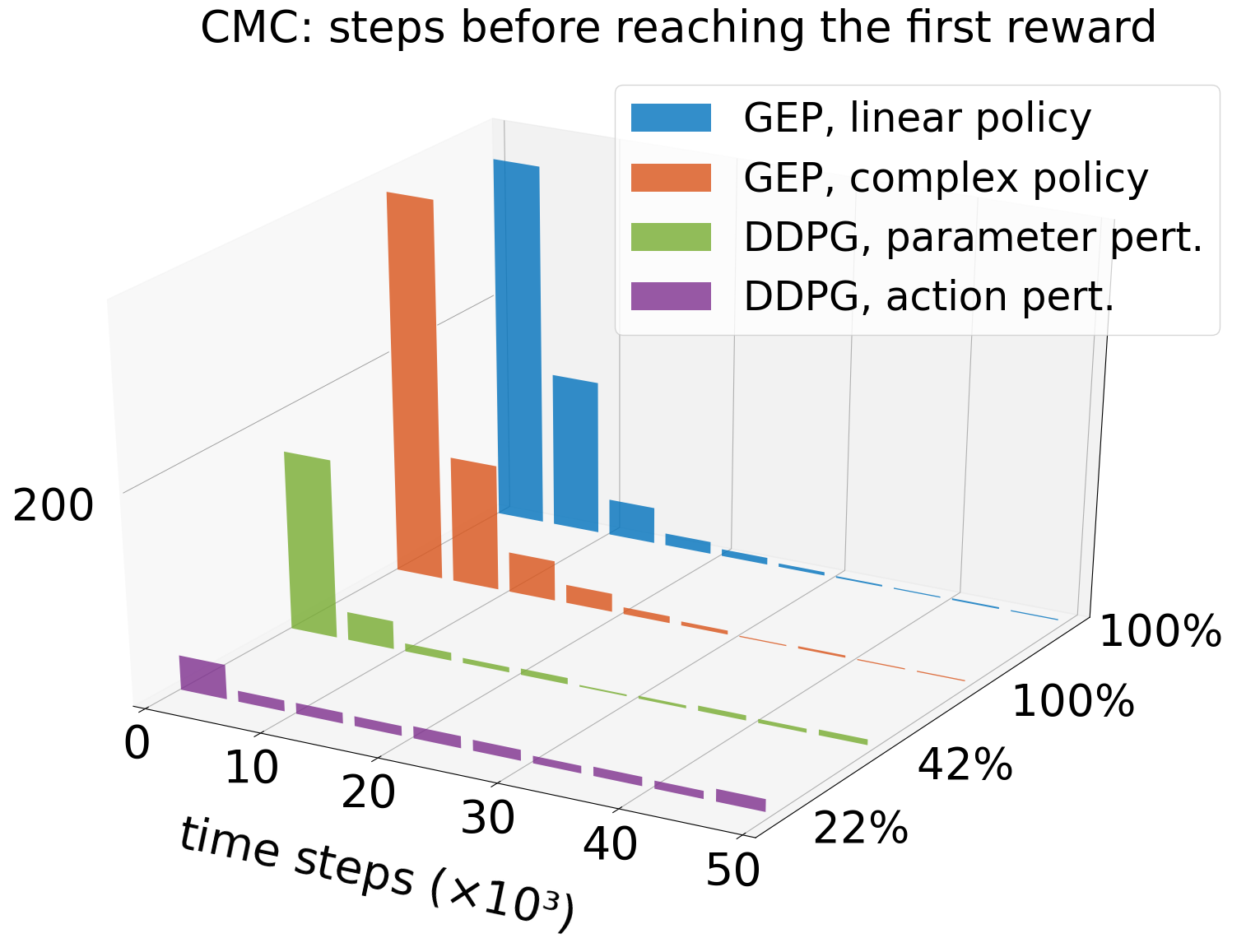}
  \caption{Histogram of number of steps to reach the goal for the first time with: 1)\gep applied to the linear policy; 2) \gep applied to a deep policy; 3) \ddpg with noise on the network parameters and 4) \ddpg with OU noise $(\mu=0, \sigma=0.3)$ on actions\label{fig:histo}. The percentages of times the goal was found before $5.10^4$ steps is indicated on the y-axis.}
\end{figure}

Figure~\ref{fig:histo} represents the histograms of the number of steps before the goal is reached for the first time on CMC.
We compare four algorithms: \gep applied to the linear policy we use in \geppg, \gep applied to the same policy as \ddpg $(64,64)$ that we call {\em complex policy} thereafter, \ddpg with noise on the network parameters and \ddpg with OU noise $(\mu=0, \sigma=0.3)$ on actions. We also tried \gep applied to a policy with hidden layers $(400,300)$, but the result is not statistically different from the one obtained with the $(64,64)$ policy.

For each condition, we perform $1000$ trials and we record the mean and the histogram of the number of steps necessary to reach the goal.
However, because both variants of \ddpg might never reach the goal, we stop experiments with these algorithms if they have not reached the goal after $5.10^4$ steps, which is more than the maximum number of steps needed with \gep variants.

The \gep algorithm using a simple linear policy and the complex policy take on average $3875$ and $3773$ steps respectively to find the top.
Using \ddpg with parameter perturbation (respectively action perturbation), the goal is reached before $5.10^4$ steps only in $42\%$ (respectively $22\%$) of the trials.
These results are enough to show that \gep finds the goal faster than the \ddpg variants.
This effect can be attributed to the deceptive gradient issue present in CMC, as performing SGD before finding the goal drives policy improvement into a wrong direction. 

The \gep variants actually reach the goal as early as the bootstrap phase where the policy is essentially random. Thus, despite the deceptive gradient effect, random exploration is enough to solve CMC. This confirms the idea that CMC could be considered a simple problem when the algorithm is not blinded by the reward feedback.

Finally, the policy complexity of \gep does not seem important. What matters is the ratio between the size of $\Theta$ and the subspace of successful policy parameters, rather than the size of $\Theta$ itself.

\begin{figure*}[hbpt]
  \centering
 \subfigure[\label{perf_cmc}]{\includegraphics[width=0.65\columnwidth]{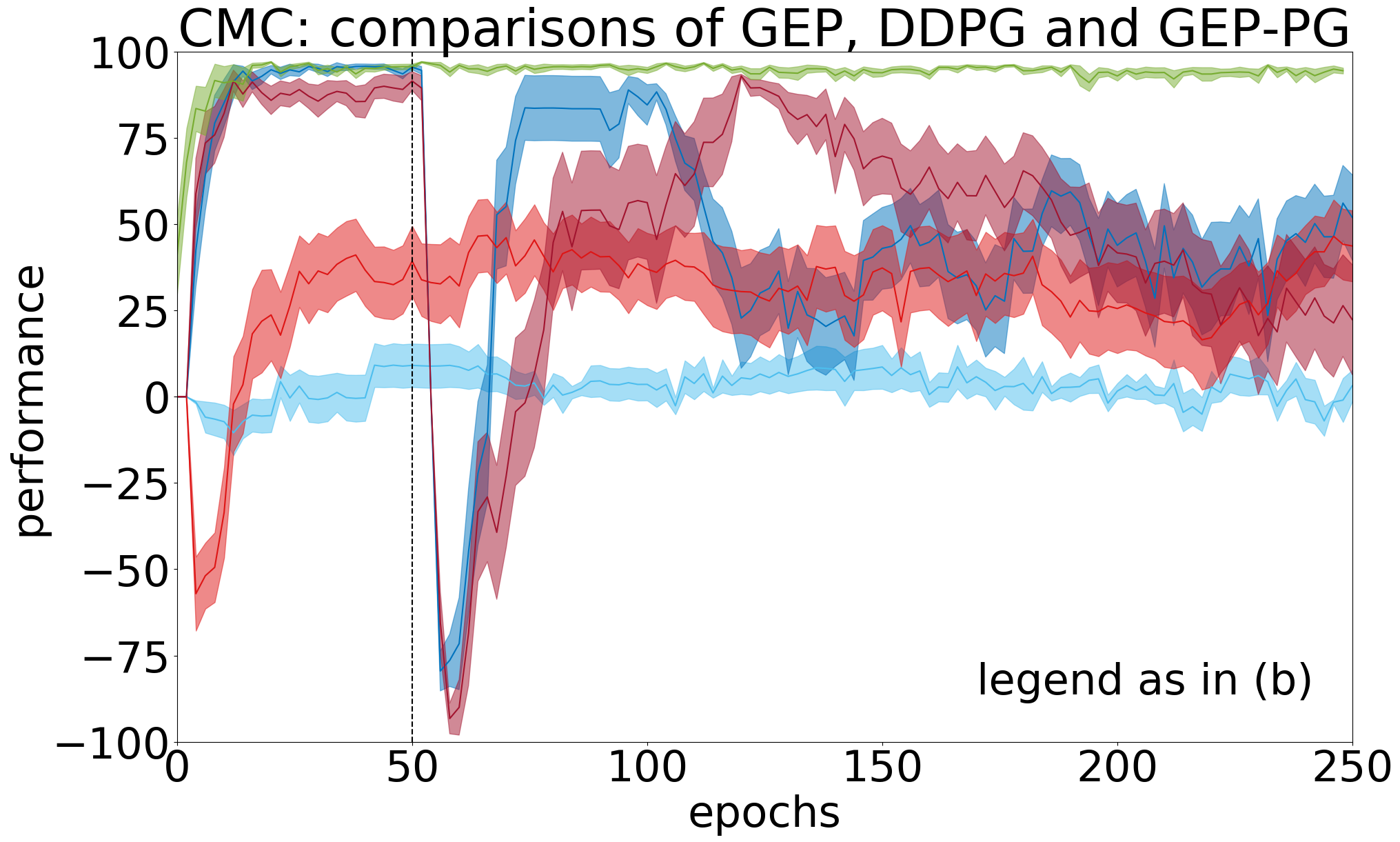}}
 \subfigure[\label{histo_cmc}]{\includegraphics[width=0.55\columnwidth]{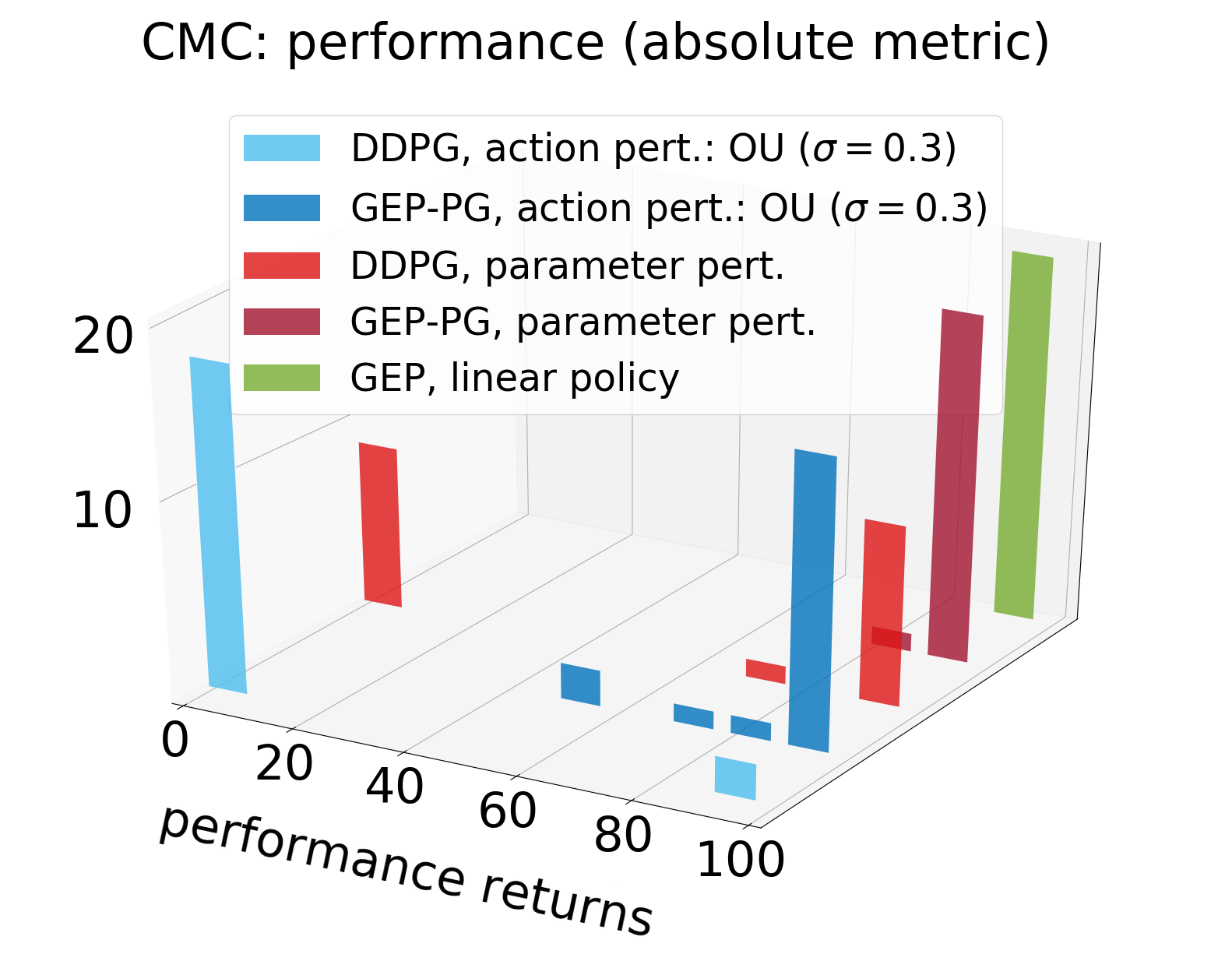}}
 \subfigure[\label{perf_hc}]{\includegraphics[width=0.65\columnwidth]{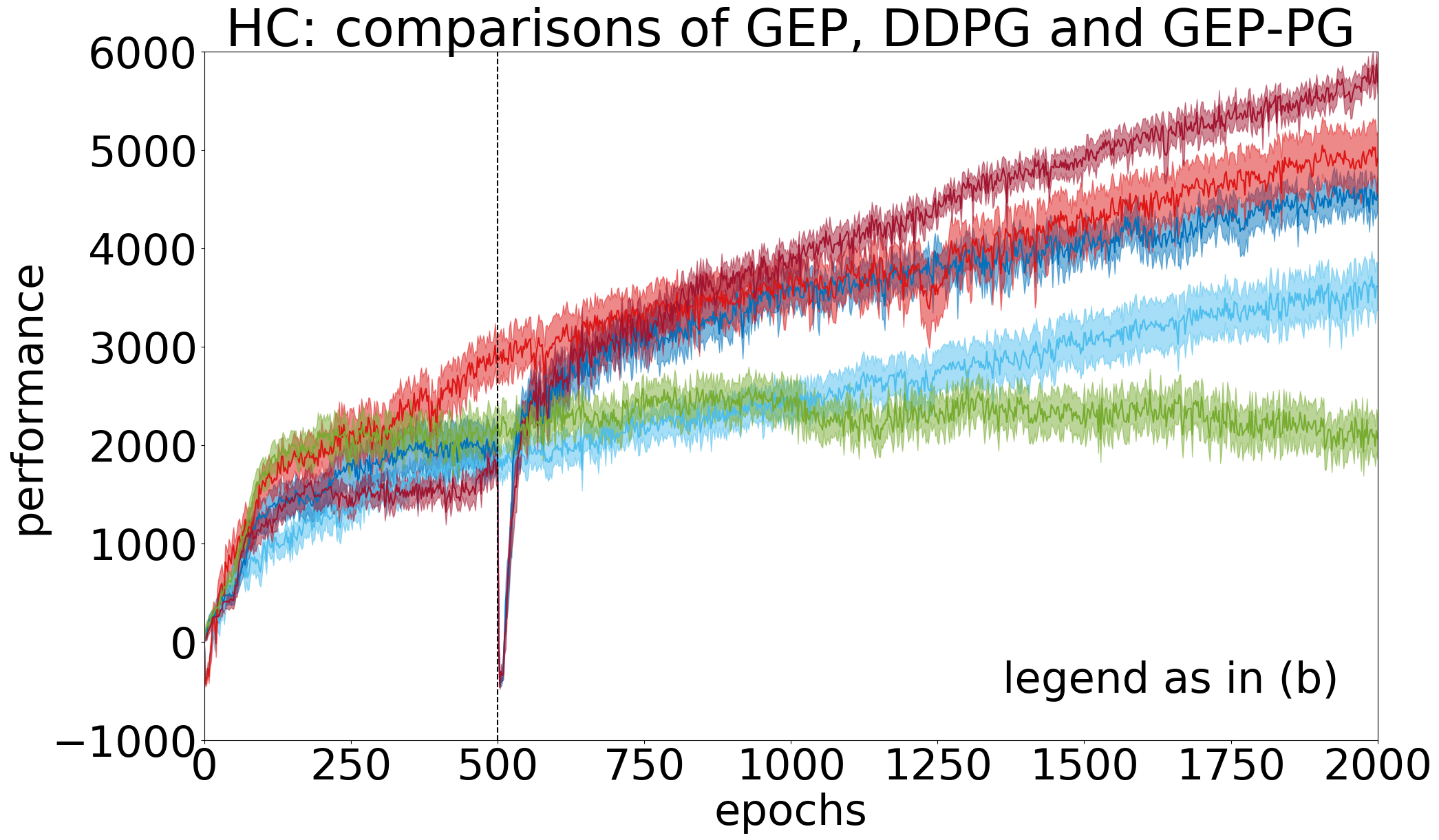}}
 \caption{(a): Learning performance on CMC of \gep, \ddpg with OU noise $(\mu=0, \sigma=0.3)$ on actions, \ddpg with noise on policy parameters and \geppg with the same noise configurations. (b): Histogram of absolute performances on CMC (c): Same as (a) for \hc. \label{fig:perf}}
\end{figure*}

\subsection{Combining \gep and \ddpg into \geppg}

In this section, we compare the performance of \ddpg with action perturbation and with parameter perturbation to the corresponding \geppg versions and to \gep alone.
In ~{\figurename~\ref{perf_hc}}, the vertical dotted line signals the transition from the \gep part of \geppg to the \ddpg part. In \figurename~\ref{perf_cmc}, the dotted line is indicative, because \geps perform $50$ episodes and 1 episode does not always last $10^3$ steps in CMC.

In \geppg performance curves, performance drops when switching from \gep to \ddpg because the policy is not transfered from one algorithm to the other (only samples gathered by \gep are transmitted to the replay buffer of \ddpg). In the case where \gep is using a \ddpg-like policy, such transfer would be possible, but this is left for future work.

As in Section~\ref{sec:undirec}, the learning performance depicted in \figurename~\ref{perf_cmc} does reflect the ratio of trials which found the goal rather than an average performance and these ratios are better depicted in the histogram of \figurename~\ref{histo_cmc}. Nevertheless,
Figure~\ref{perf_cmc} shows that \gep alone quickly finds an efficient policy and remains stable at a good performance. This stability is easily explained by the fact that, being a lazy learning approach, a \gep cannot forget a good solution to a stationary problem, in contrast with \ddpg. We can also see that \ddpg with standard OU noise performs surprisingly poorly on this low dimensional benchmark, and that \ddpg with noise on the network parameters does not perform much better. Finally, the \geppg counterparts of the \ddpg variants perform slightly better, but not at the level of \gep.

From these results, \gep alone being better, one may wonder whether \geppg and deep RL approaches in general are useful at all. The point that they are will be made when using the higher dimensional \hc benchmark.

On \hc, \gep alone using the linear policy performs statistically worse than others on both performance metrics, a performance that is not due to the simplicity of its policy since using the $(64,64)$ neurons policy provides results that are not significantly different (see Appendix F).

More importantly, the \geppg versions  significantly outperform their \ddpg counterparts on both metrics and reach a new \sota performance on this benchmark where \ddpg with noise on the network parameters was the current leader \citep{henderson2017deep}. The performance of final policies and best policies found over the whole learning process are given in Table~\ref{tab:perf}. Note that learning curves of \ddpg without exploration noise and its \geppg counterpart are not represented here, as they were found to match closely the corresponding versions using OU(0.3) noise.

\begin{table}[!htb]
  \centering
  \caption{Final and absolute performances on \hc, mean (std)\label{tab:perf}}
  \begin{tabular}{ccc}
    \hline
    Algorithm & final metric & absolute metric\\
    \hline
    \ddpg action pert. & 3596 (1102) & 3943 (1209) \\
    \ddpg param. pert. & 4961 (1288) & 5445 (1265)\\
    \geppg action pert. & 4521 (738) &  5033 (833)\\
    \geppg param pert. & {\bf 5740 (573)} & {\bf 6118 (634)} \\
    \gep param pert. & 2105 (881) & 2140 (881) \\
    \hline
  \end{tabular}
\end{table}

\geppg is found to be robust to the number of \gep episodes used to fill the \ddpg replay buffer. Performance is stable from about 100 to 600 \gep episodes (see Appendix G).
Finally, the \geppg versions seem to generate less variability than their \ddpg counterparts, which means that an efficient policy can be found more consistently (see standard deviations in Table \ref{tab:perf} and Appendix E, Figure 6). Performance predictability matters if performance should not fall below a specific level (e.g. for safety reasons).

From \figurename~\ref{fig:perf}, one may consider that, on both CMC and \hc, \geppg with policy parameter noise outperforms \geppg with standard OU noise on both metrics. Further analyses of the impact of the content of a buffer filled with \gep on \geppg performance led to the following conclusions: 1) the size of the buffer does not impact \geppg performance (from $100$ to $2.10^3$ episodes); 2) \geppg performance correlates with \gep best performance and the average performance of training policies; 3) \geppg performance correlates to diversity in terms of observations and outcomes as measured by various metrics (see Appendix G). Therefore, a good buffer should contain both efficient and diverse trajectories.

Note that, as it is used here, \geppg only uses 0.3\% additional storage and has a higher complexity  than \ddpg w.r.t the number of episodes: $\mathcal{O}(n^2\log{}n)$ vs. $\mathcal{O}(n)$. However, due to complexity constants, it is much faster in practice for the number of episodes we use.

\section{Discussion and conclusion}
\label{sec:conclusion}

Reinforcement learning problems with sparse or deceptive rewards are challenging for continuous actions algorithms using SGD, such as \ddpg.
In sparse reward problems, the necessary gradient to perform policy improvement may not be available until a source of reward is found. Even worse, in deceptive reward
problems, the gradient may drive to wrong directions. The baseline version of \ddpg, using random action perturbation is highly inefficient in these contexts as shown on the CMC and \hc benchmarks. We have reproduced results showing that policy parameter noise is a better strategy, but our experiments show that this form of exploration is still insufficient. 
So how should we organize exploration, especially in the early stages of learning, to be robust to rare and deceptive rewards?

\textbf{Decoupling exploration and exploitation}. 
We have proposed \geppg as a two-phase approach, where a first exploration phase discovers a repertoire of simple policies maximizing behavioral diversity, ignoring the reward function. In particular, we used the developmental \gep approach and analyzed its exploration efficiency on CMC. While experiments presented in this paper show that \gep alone is already competitive, it is less efficient as it gets closer to the optimal policy in larger benchmarks such as \hc. This is why, in \geppg, the exploration phase is followed by a more standard deep RL phase for fine-tuning, where \ddpg uses a replay buffer filled with samples generated by \gep.

Using two benchmarks, we have shown that training \ddpg after filling its replay buffer with a \gep 1) is more sample efficient, 2) leads to better absolute final performance and 3) has less variance than training it from scratch using standard action or parameter perturbation methods. Furthermore, we have shown that using a \gep alone was the method of choice for the simple CMC benchmark, partly due to its deceptive reward signal effect, but \geppg takes the lead and provides performance beyond the \ddpg on the larger \hc benchmark. While this paper focused on the detailed study of these algorithms in two benchmarks, future work will evaluate them more broadly. 

\textbf{Limits and future work.} We saw that \geppg was robust to the number of episodes in the initial exploration phase. A next stage could be to extend \geppg towards an adaptive transition or mixing mechanism, where a multi-armed bandit could be used to dynamically switch between both learning strategies using measures like learning progress \citep{lopes2012strategic}. 

It would be interesting to compare other variants of \geppg. For the exploration phase, more advanced forms of \gep algorithms could be used, for example using curiosity-driven sampling of goals \citep{baranes_active_2013} or closely related exploration algorithms in the evolutionary computation literature like Novelty Search \citep{conti2017improving} or Quality-Diversity \citep{cully2017quality}. For the exploitation phase, \ddpg could be replaced by the more recent \acktr \citep{wu2017scalable} which can in principle be trained off-policy. The \geppg approach could also be applied to discrete action RL problems with rare rewards, bootstrapping exploration to initialize the replay buffer of the \dqn algorithm or its variants. On the contrary, this approach can not be used with an on-policy deep RL algorithm such as \trpo or \ppo because they do not use a pivotal replay buffer.

As the \gep is intrinsically a multi-goal exploration algorithm, another line of investigation would be to study its impact for bootstrapping goal-parameterized deep RL approaches for continuous actions RL problems like Hindsight Experience Replay \cite{andrychowicz2017hindsight,held2017automatic}. This would be interesting as goal-parameterized deep RL approaches sample goals stochastically as part of their exploration mechanism (like \gep), but differ from \gep approaches as they learn a single large policy as opposed to a population of simpler policies. 

Finally, evolutionary and developmental exploration methods like \gep, Novelty Search or Quality-Diversity take advantage of a behavioral features space, which in most works characterizes properties of the state or state-action trajectories. Many instances of these algorithms have successfully used features that are relatively straightforward to define manually, but an open question is how they could be learned through unsupervised learning (see \citep{pere18} for a potential approach) and how such learned features would impact the dynamics of the \geppg approach.

{\scalefont{0.96}

\newpage
\section{Acknowledgments}
We thank Pierre Manceron for his strong investment in an initial version of this work and Pierre Fournier for useful discussions. This research is financially supported by the French Minist\`ere des Arm\'ees - Direction G\'en\'erale de l'Armement.
This work was supported by the European Commission, within the DREAM project, and has received funding from the European Unions Horizon 2020 research and innovation program under grant agreement $N^o$ 640891.


}

\newpage
\appendix
\icmltitle{Appendices}
\section{Study of \ddpg variability}
In this section, we investigate performance variability in \ddpg. The authors of \cite{henderson2017deep} showed that averaging the performance of two randomly selected splits of 5 runs with different random seeds can lead to statistically different distributions. This considerably undermines previous results, as most of the community (\cite{henderson2017deep} included) has been using 5 seeds or less, see \cite{henderson2017deep}. 
Here we use a larger statistical sample of 20 random seeds and show that it is enough to counter the variance effect. We run the baseline \ddpg algorithm with OU($\sigma=0.3$) noise 40 times on Continuous Mountain Car and Half-Cheetah. We randomly select two pairs of $20$ sets and perform statistical tests to compare their performance. We repeat this procedure $1000$ times and report the percentages of tests showing a significant difference between both sets (Table~\ref{tab:t1} and \ref{tab:t2}). We report results for both evaluation metrics and two tests: two-sample t-test (\ttest) and bootstrapped (\bs) estimation of the 95\% confidence interval for the mean difference (positive if the interval does not contain $0$).

 \begin{table}[!htb]
  \centering
    \caption{Percentage of tests showing differences between two sets of the same algorithm: \ddpg OU(0.3) on (a) CMC, (b) \hc. \label{tab:variability}}

\subfigure[\label{tab:t2}]{
  \begin{tabular}{|ccc|}
  
    \hline
    &abs. metric & final metric\\
    \hline
    \ttest & 0.0\% & 0.0\%\\
    \bs & 0.0\% & 5.1\% \\
    \hline
  \end{tabular}
  }
   \subfigure[\label{tab:t1}]{
  \begin{tabular}{|ccc|}
  
    \hline
    &abs. metric & final metric\\
    \hline
    \ttest & 0.0\% & 0.0\% \\
    \bs & 0.0\% & 0.0\% \\
    \hline

  \end{tabular}
}
  \end{table}

\section{Correlation between evaluation metrics}

We use two performance metrics: (1) the {\em absolute metric} is the average performance over 100 test episodes using the best controller over the whole learning process; (2) the {\em final metric}, corresponding to the evaluation methodology of \cite{henderson2017deep} is the average over the $100$ last test episodes of the learning process, $10$ episodes for each of the last $10$ different controllers. 

\begin{figure}[t]
  \centering
 \subfigure[\label{noise_cmc_a}]{\includegraphics[width=0.9\linewidth]{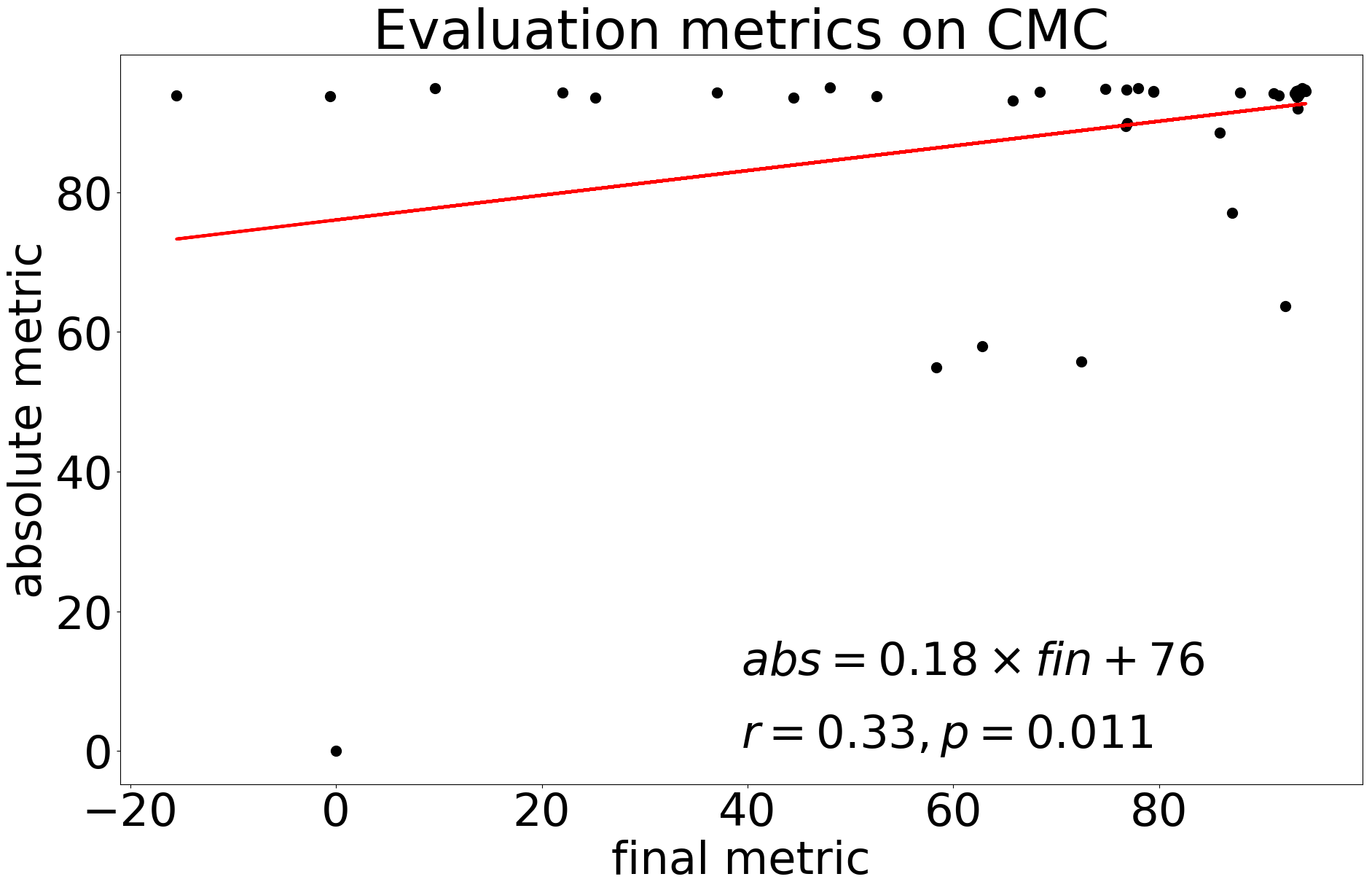}}
 \subfigure[\label{noise_hc_a}]{\includegraphics[width=0.9\linewidth]{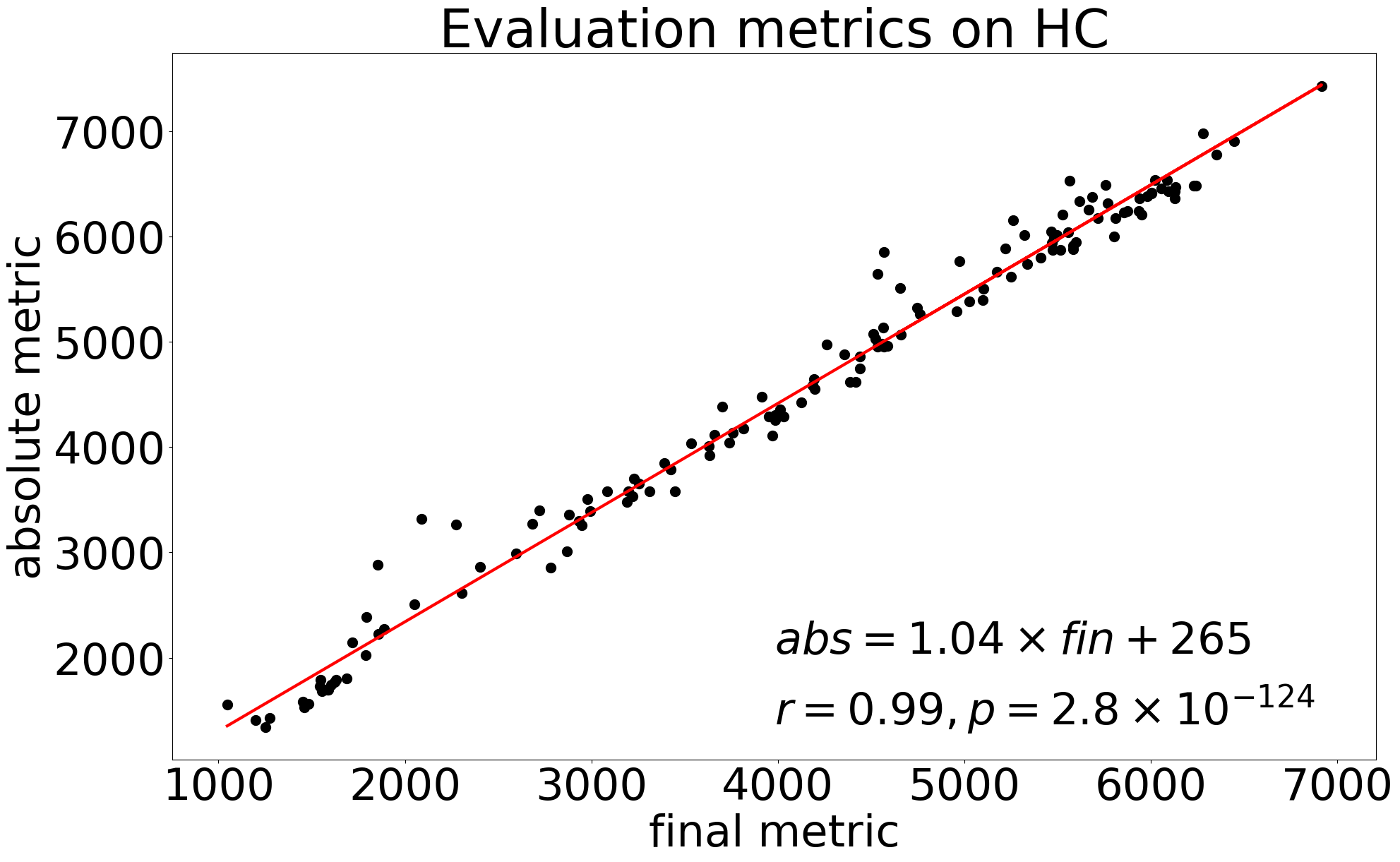}}
 \caption{Correlation of the two metrics of performance on (a) CMC and (b) \hc. The equations of the lines of best fit (in red), the Pearson coefficients $r$ and their associated p-values $p$ are provided. \label{fig:correl}}
\end{figure}

As highlighted in Appendix-Figure~\ref{fig:indiv} of Appendix~\ref{sec:indiv}, \ddpg performance on CMC is highly unstable. The final metric only gives a measure of the final performance, which might not represent the algorithm's ability to find a good policy. Figure~\ref{noise_cmc} shows this problem for the \geppg runs: even though the Pearson correlation coefficient is found significant, the line slope is far from 1. The final metric is highly variable whereas the absolute metric almost always shows a good performance. On the opposite, Appendix Figure~\ref{noise_hc} shows that final metrics and absolute metrics of \ddpg performance on \hc are highly correlated with a slope close to 1. This can be seen on Figures~1(c) and 2(c) where all learning curves are strictly increasing: the highest performance is always among the last ones. As a result, it is better to report the {\em absolute metric}, representing the performance of the best policy over a run. In the case of unstable learning as in CMC, we find it informative to present different runs so as to get a better sense of the learning dynamics. This is done in the next section.
  
\section{Individual runs on CMC}
\label{sec:indiv}

\begin{figure}[H] 
    \centering
    \includegraphics[width=1\linewidth]{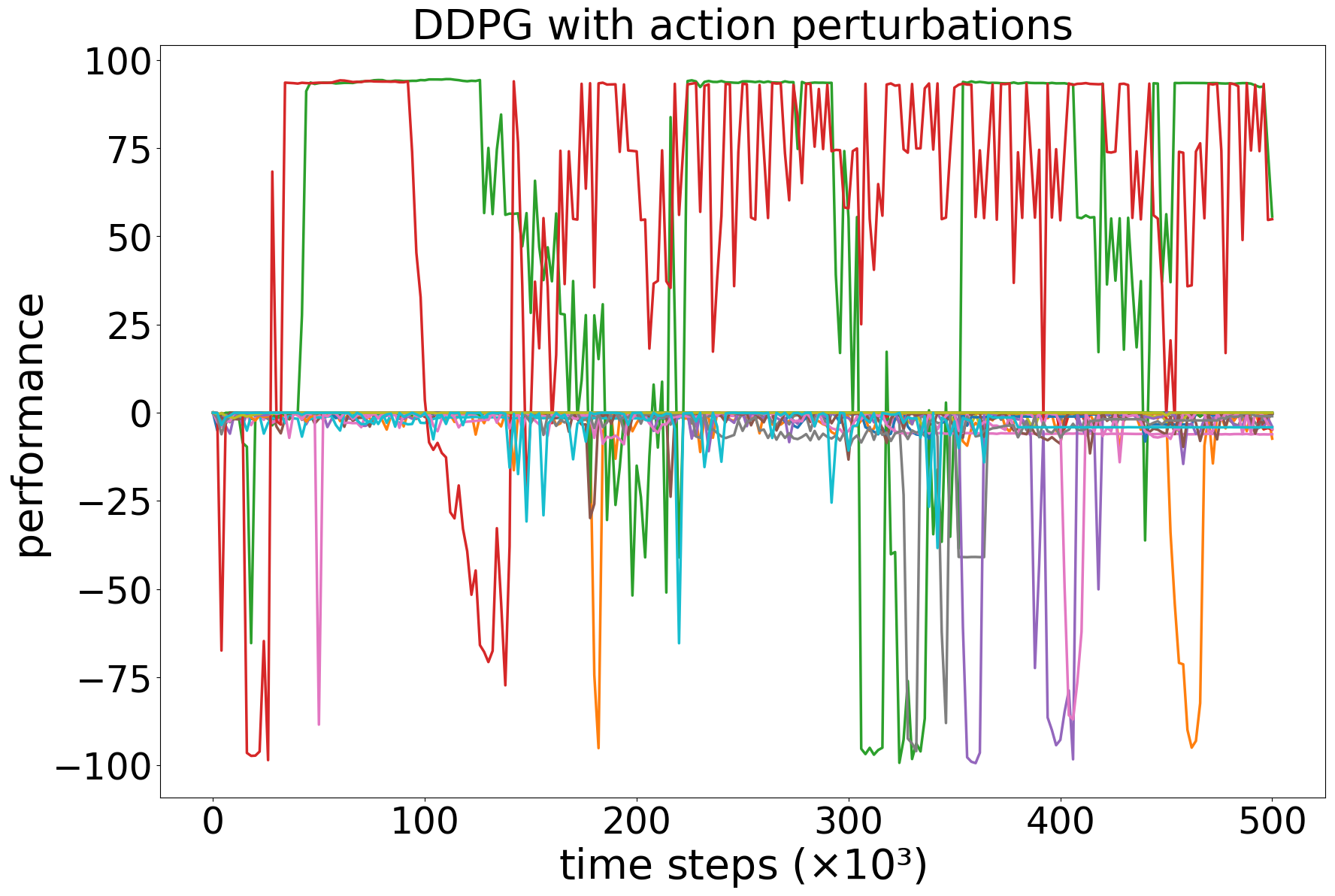} 
    \caption{Twenty individual runs of \ddpg OU(0.3) on CMC\label{fig:indiv}} 
\end{figure}

Figure~\ref{fig:indiv} shows a representative example of 20 runs of \ddpg with standard OU noise $(\mu=0,\sigma=0.3)$. One can see that most runs never find the rewarding goal and that the learning performance of those which do so is unstable. As a result, the distribution of performances is not normal. 

\section{Performance comparisons}

Below we present statistical comparisons of the performance of various pairs of algorithms on \hc. For comparisons, we use the 2-sample \ttest (\ttest) and a measure of the $95\%$ confidence interval computed by bootstrap (\bs) using $10^4$ samples. The scipy implementation is used for \ttest and the Facebook Bootstrapped implementation for \bs \footnote{https://github.com/facebookincubator/bootstrapped}. For the \ttest, we present the test value with the p-value in brackets, the difference being significant when $p \leq 0.05$. For \bs, we present the mean and bounds of the $95\%$ confidence interval for the difference between the two sets of performances. It is positive if the interval does not contain $0$.

\begin{table}[H]
  \centering
  \caption{\ddpg param. pert. versus \ddpg action pert. }
  \begin{tabular}{|ccc|}
    \hline
    &absolute metric & final metric\\
    \hline
    \ttest & 3.74 (6.0$\times10^{-4}$) & 3.5 (1.2$\times10^{-3}$)\\
    \bs & 1502 (736, 2272) & 1364 (641, 2120)\\
    \hline
  \end{tabular}
\end{table}

\ddpg with parameter perturbation achieves a significantly higher final (2/2) and absolute performance (2/2) than \ddpg with action perturbation.

\begin{table}[H]
  \centering
  \caption{\ddpg OU(0.3) versus decreasing OU(0.6), action pert. }
  \begin{tabular}{|ccc|}
    \hline
    &absolute metric & final metric\\
    \hline
    \ttest & 1.1 (0.29) & 1.4 (0.16)\\
    \bs & 425 (-330, 1196) & 534 (-151, 1257)\\
    \hline
  \end{tabular}
\end{table}

There is no statistical evidence that \ddpg with OU(0.3) achieves higher final or absolute performance than the decreasing OU(0.6) version. 

\begin{table}[H]
  \centering
  \caption{\ddpg action pert. versus \gep}
  \begin{tabular}{|ccc|}
    \hline
    &absolute metric & final metric\\
    \hline
    \ttest & 5.3 (7.6$\times10^{-6}$)& 4.6 (4.9$\times10^{-5}$)\\
    \bs & 1804 (1151, 2476) & 1491 (893, 2115)\\
    \hline
  \end{tabular}
\end{table}
\ddpg with action perturbation achieves significantly higher final and absolute performance than \gep, 2/2 tests for both metrics.

\begin{table}[H]
  \centering
  \caption{\gep linear policy versus \gep complex policy (64,64)}
  \begin{tabular}{|ccc|}
    \hline
    &absolute metric & final metric\\
    \hline
    \ks & 0.15 (0.96) & 0.20 (0.77) \\
    \ttest & 0.32 (0.75)& 0.52 (0.60)\\
    \bs & 787 (-404, 552) & 128 (-350, 595)\\
    \hline
  \end{tabular}
\end{table}

The complexity of the \gep policy is not found to make any difference in absolute or final performance.

\begin{table}[H]
  \centering
  \caption{\geppg versus \ddpg with action perturbation}
  \begin{tabular}{|ccc|}
    \hline
    &absolute metric & final metric\\
    \hline
    \ttest & 3.2 (2.7$\times10^{-3}$)& 3.0 (4.6$\times10^{-3}$)\\
    \bs & 1089 (448, 1726) & 924 (334, 1503) \\
    \hline
  \end{tabular}
\end{table}

\geppg achieves significantly higher absolute and final performance than \ddpg, both using action perturbation (2/2 tests for both performance metrics)

\begin{table}[H]
  \centering
  \caption{\geppg versus \ddpg, parameter perturbation}
  \begin{tabular}{|ccc|}
    \hline
    &absolute metric & final metric\\
    \hline
    \ttest & 2.1 (4.8$\times10^{-2}$)& 2.4 (2.3$\times10^{-2}$)\\
    \bs & 672 (39, 1261) & 780 (143, 1378)\\
    \hline
  \end{tabular}
\end{table}

\geppg achieves significantly higher absolute and final performance than \ddpg, both using parameter perturbation (2/2 tests with both performance metrics).

\section{Histograms of performance on \hc}
\begin{figure}[H]
  \centering
 \subfigure[\label{hist_cmc}]{\includegraphics[width=0.9\linewidth]{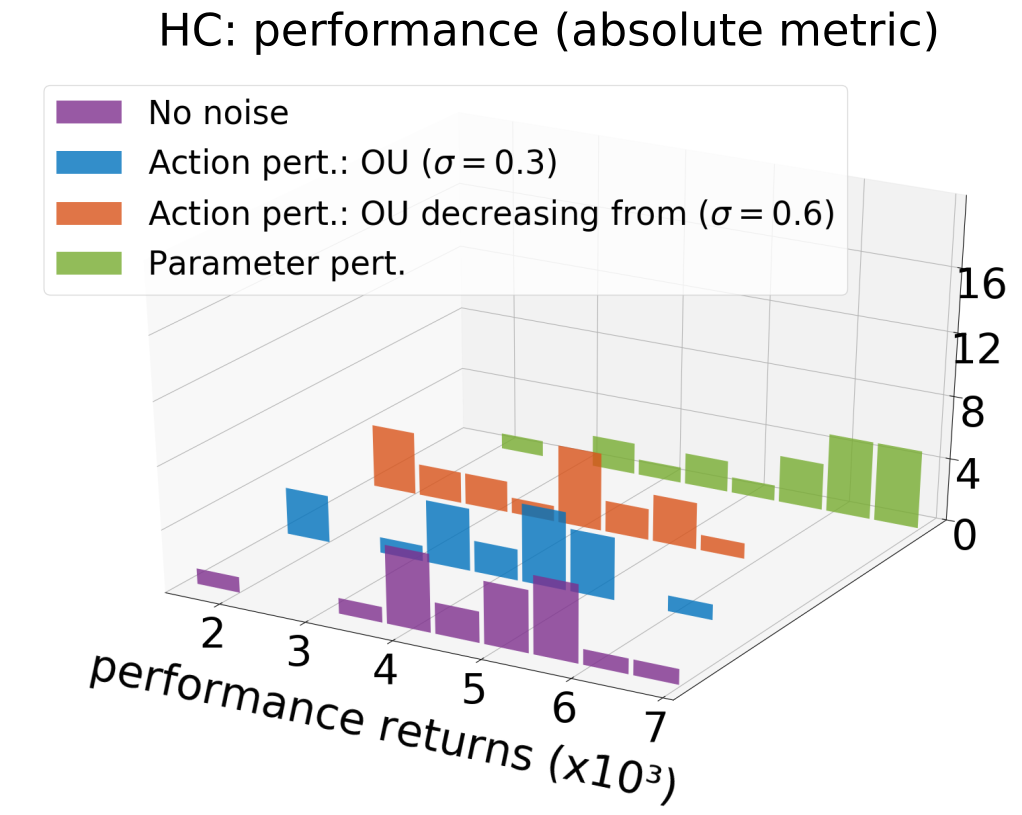}}
 \subfigure[\label{hist_hc}]{\includegraphics[width=0.9\linewidth]{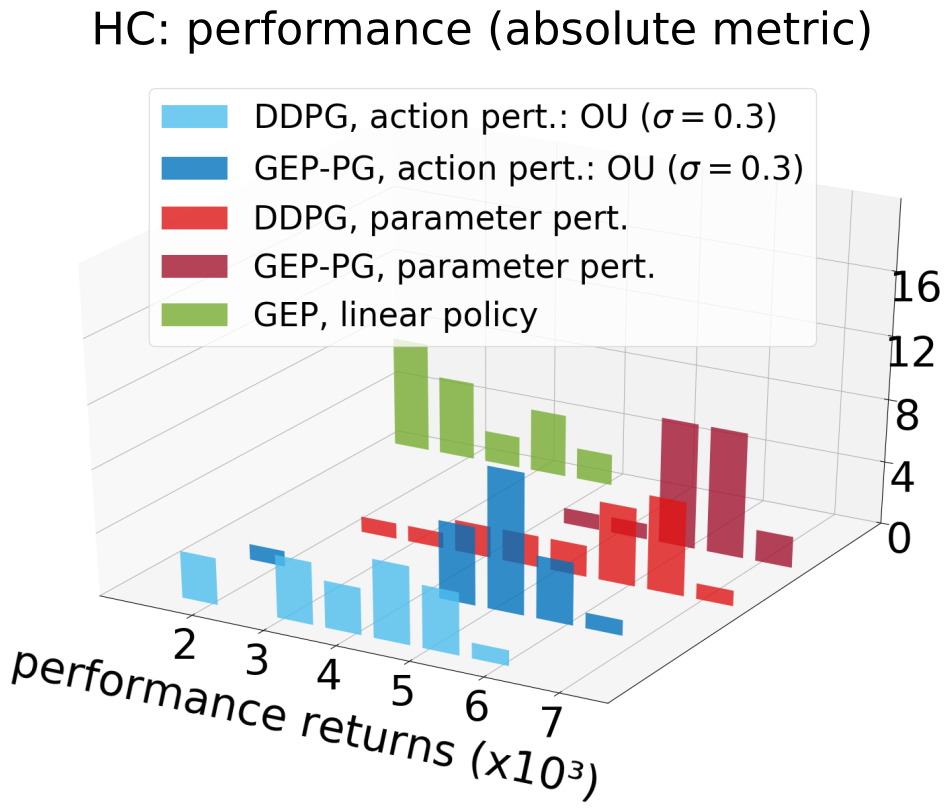}}
 \caption{Absolute metric in \hc for the various algorithms.(a) shows the influence of undirected exploration with perturbations on performance. (b) show the performances of  \gep, \ddpg and their combination \geppg on \hc. \label{fig:hists}}
\end{figure}

Here we show the histograms of the absolute metrics of \hc (\figurename~\ref{fig:hists}). We can see that the \geppg versions of \ddpg algorithms show a smaller variance.

\section{Influence of policy complexity in \gep}

Using a linear or a more complex policy with \gep does not impact the final \gep performances on CMC or \hc (no test over 5 shows significance). However, in the case of \hc, the version of \gep using a simple linear policy achieves higher performance sooner. The statistical tests show significance at $2.10^5$ steps with $p=7.3\times10^{-4}$ for \ks, $p=2.3\times10^-4$ for \ttest and a bootstrapped confidence interval of $738$ $(395, 1080)$. This is important in terms of sample efficiency and a \ddpg replay buffer of $2.10^5$ samples filled by \gep would probably be of higher interest if the policy was linear. This supports the idea developed in Section 3.3 that a smaller policy parameter space might be faster to explore.

\begin{figure}[H] 
   \centering
    \includegraphics[width=1.0\linewidth]{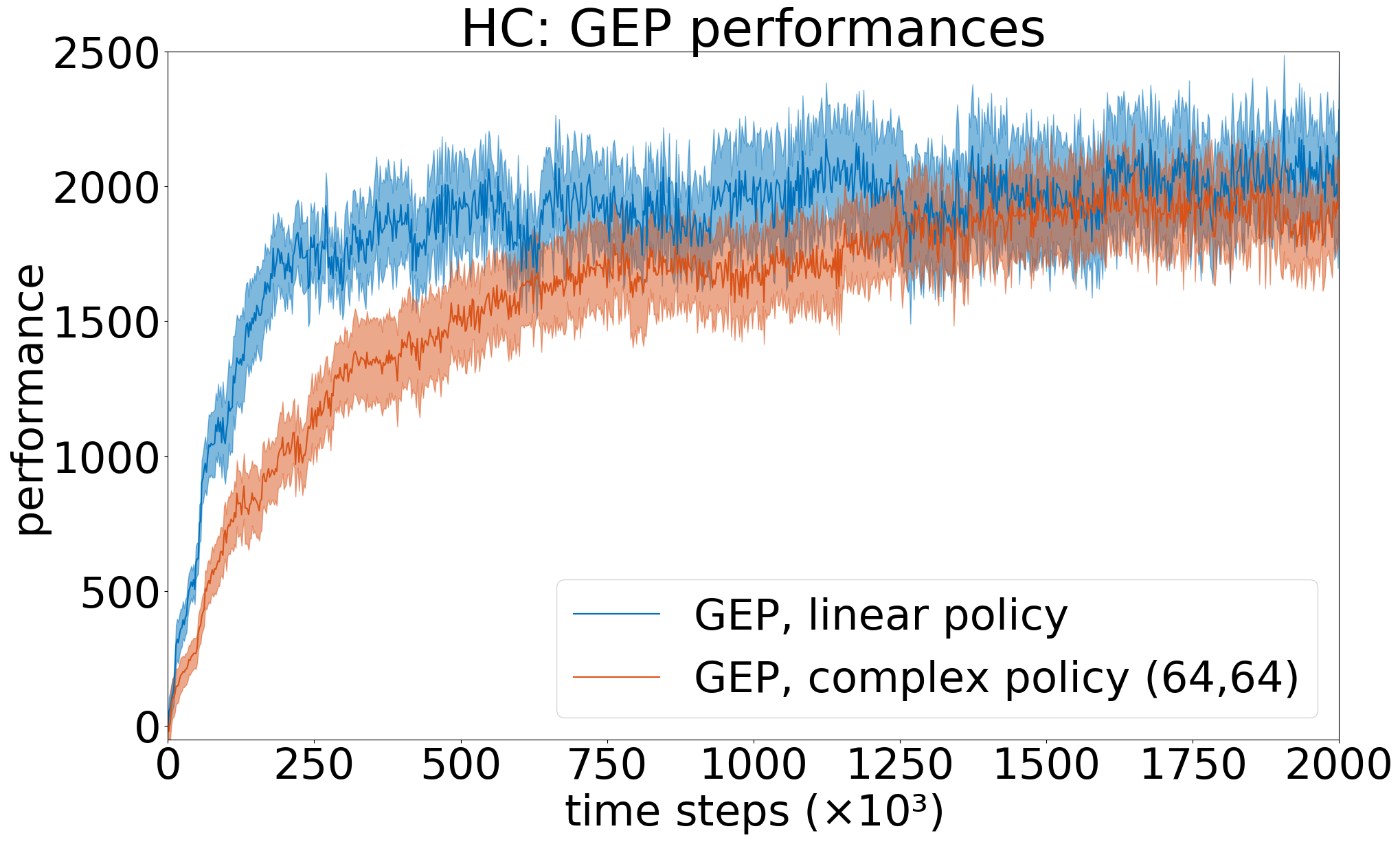} 
    \caption{Performance of \gep for a linear and a policy with two hidden layers of (64,64) neurons.\label{fig:lin_complex}}
\end{figure}

\section{Influence of the initial replay buffer content}
\begin{figure}[H] 
   \centering
    \includegraphics[width=1.0\linewidth]{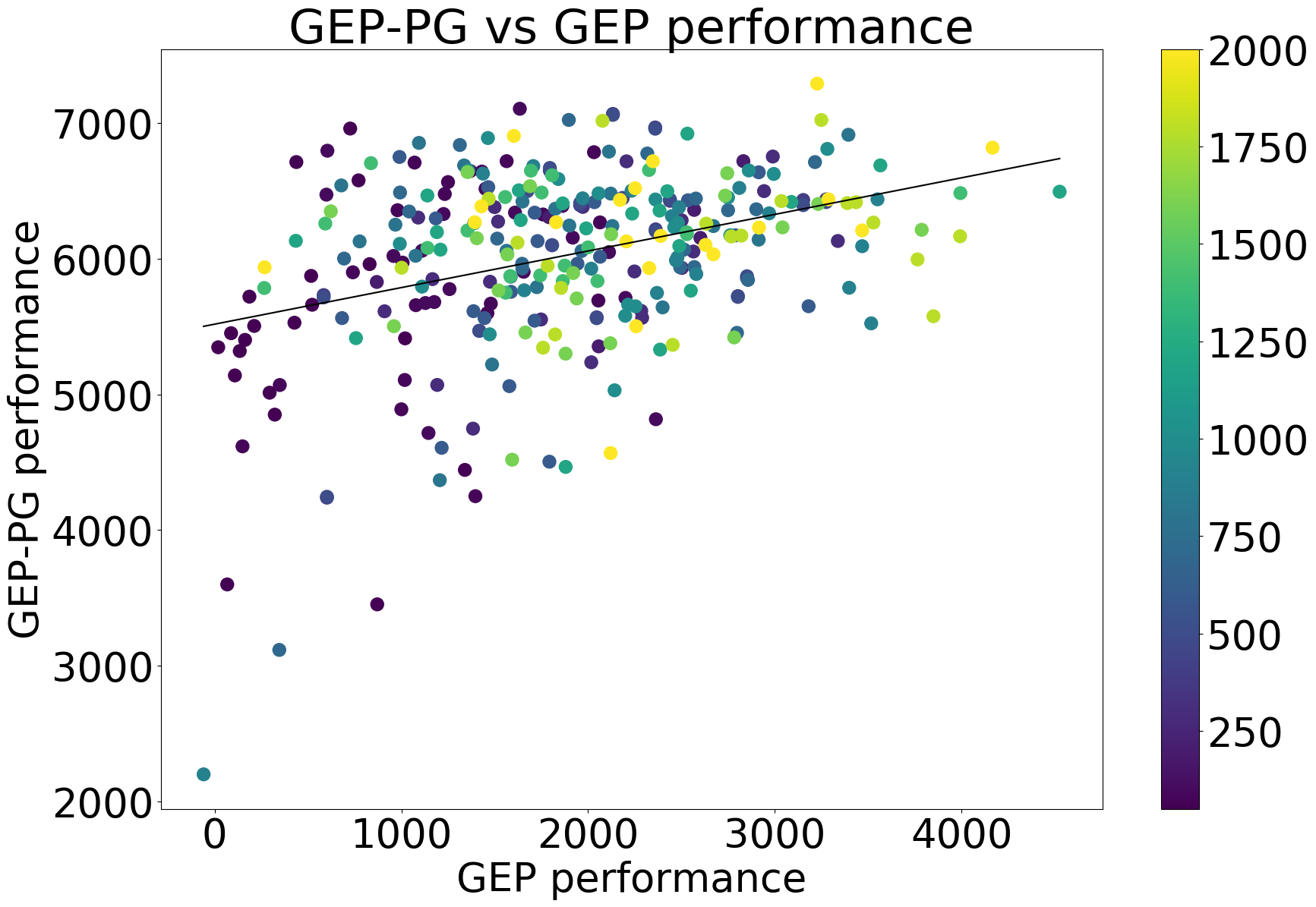} 
    \caption{\geppg performance as a function of the \gep buffer performance. The \gep performance is evaluated by replaying 100 times the best policy found by \gep. Color maps for the size of the replay buffer (the number of episodes played by \gep). \label{fig:geppg_gep}}
\end{figure}

Here we study the influence of the content of the replay buffer filled by \gep to bootstrap \ddpg on the performance of \geppg. First, we found that the size of this buffer does not influence \geppg performance (from 100 to 2000 episodes), although too few episodes harms \geppg performance ($<100$ episodes; \figurename~\ref{fig:geppg_gep}). Second, we found that \geppg performance correlates with the performance of the best \gep policy $(p<2\times10^{-6})$ (Appendix-\figurename~\ref{fig:geppg_gep}) and the average performance of all \gep policies $(p<4\times10^{-8})$. Third, we found correlations between \geppg performance and various measures of exploration diversity: 1) the standard deviation of \gep policies performances $(p<3\times10^{-10})$; 2) the standard deviation of the observation vectors averaged across dimensions. This quantifies the diversity of sensory inputs. $(p<3\times10^{-8})$; 3) the outcome diversity measured by the average distance to the k-nearest neighbors in outcome space (for various k). This measure is normalized by the average distance to the 1-nearest neighbor in the case of a uniform distribution, which makes it insensitive to the sample size $(p<4\times10^{-10})$, see Appendix-\figurename~\ref{fig:diversity}; 4) the percentage of cells filled when the outcome space is discretized (with various number of cells). We also use a number of cells equal to the number of points, which make the measure insensitive to this number $(p<4\times10^{-5})$; 5) the discretized entropy with various number of cells $(p<6\times10^{-7})$.

\begin{figure}[H] 
   \centering
    \includegraphics[width=1.0\linewidth]{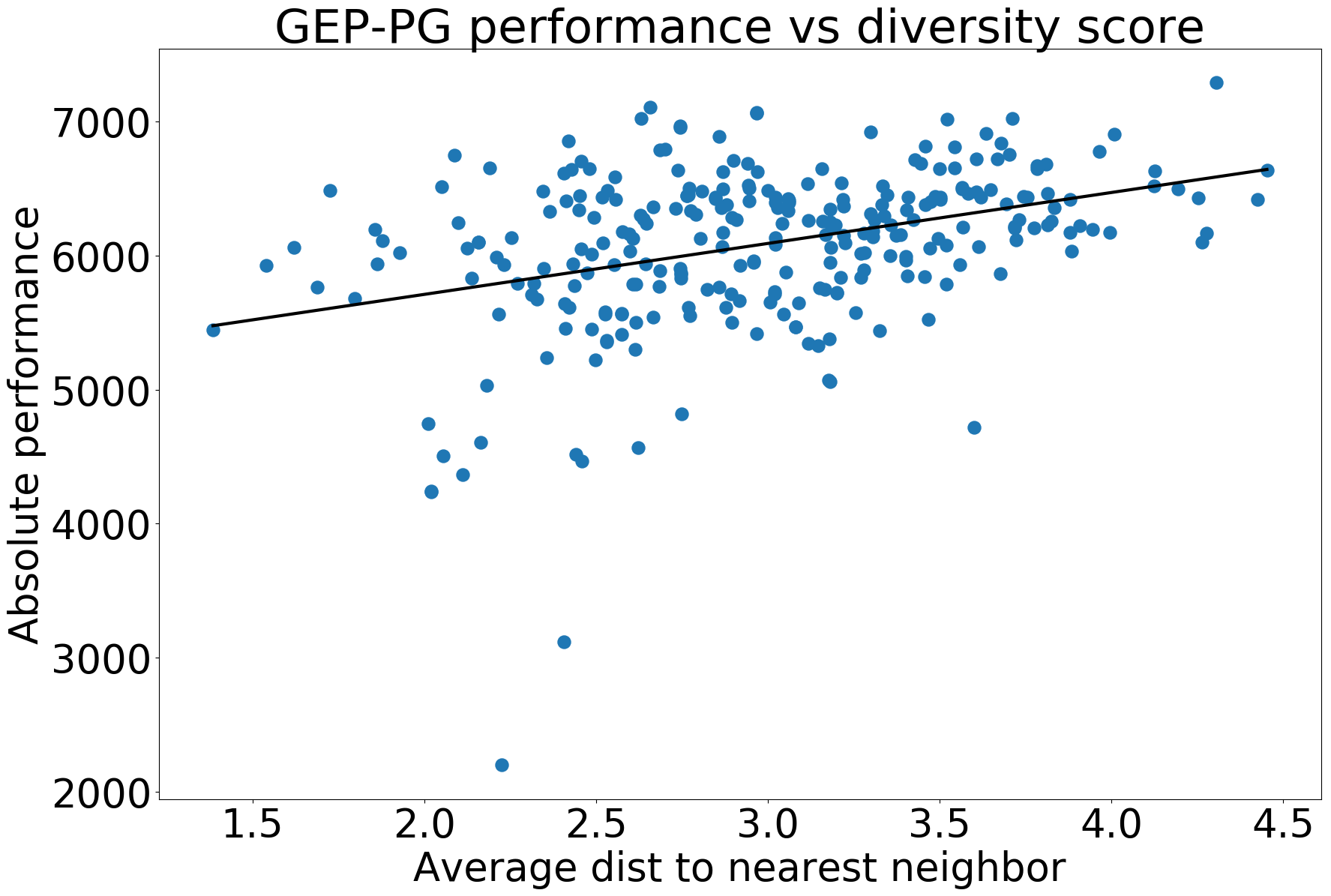} 
    \caption{\geppg performance as a function of the buffer's diversity measure. The diversity measure is computed as the average distance to the nearest neighbor in outcome space. It is normalized by the expected value of this measure in the case of a uniform distribution of outcomes. This normalization makes the measure insensitive to the number of considered samples (here through the size of the buffer). \label{fig:diversity}}
\end{figure}

\section{Sanity check}

\begin{figure}[H] 
   \centering
    \includegraphics[width=1.0\linewidth]{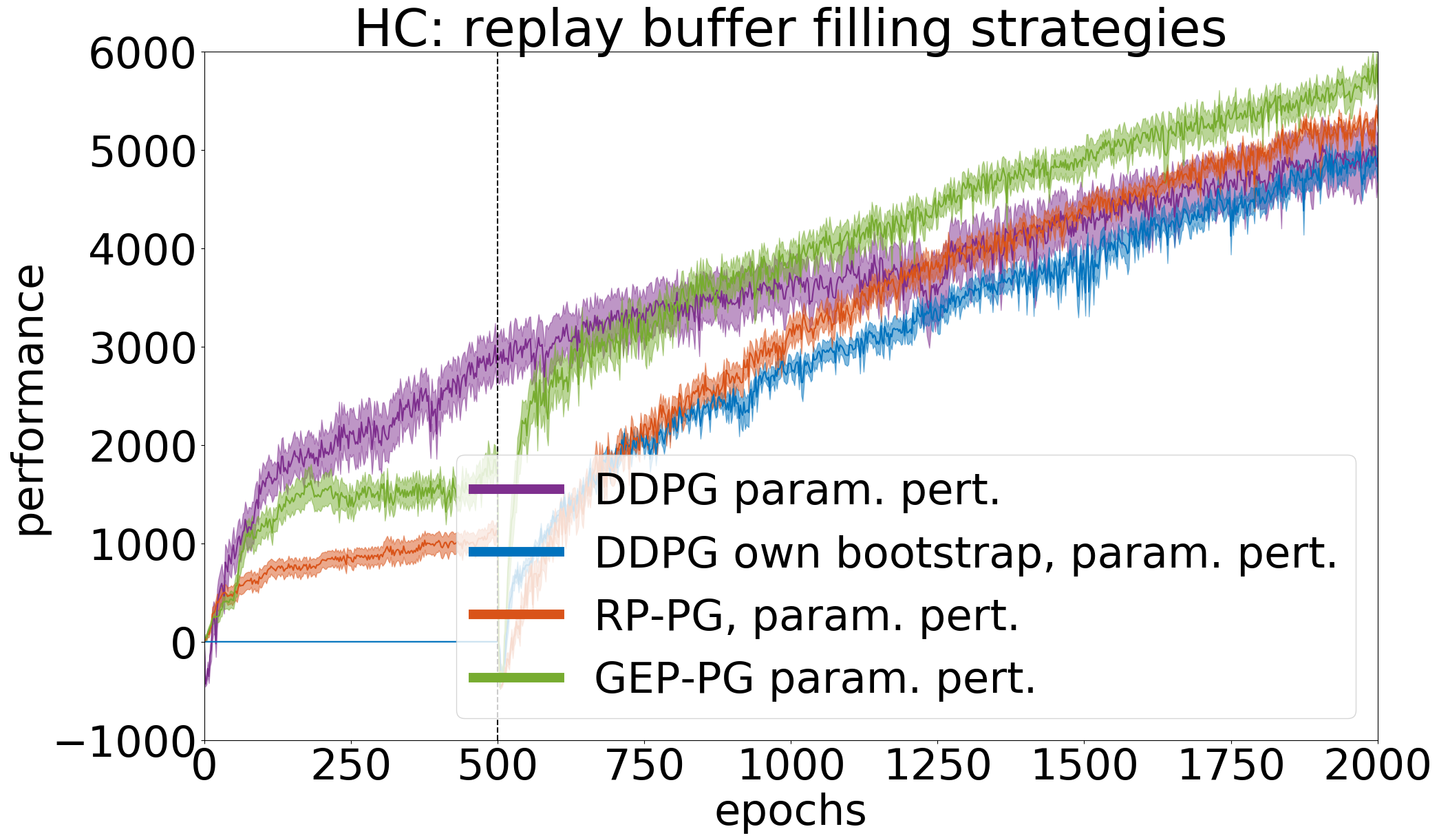} 
    \caption{Influence of various exploration strategies to initialize \ddpg's replay buffer on \geppg performance. In the second curve, the parameter perturbation exploration of \ddpg is used. In the exploration phase ($<500$ epochs), the networks are not updated. In the exploration phase of the third and fourth curves, Random Policy Search (RP) and Goal Exploration Process (GEP) are used respectively. \label{fig:sanity}}
\end{figure}

We could think of other exploration strategies to fill the replay buffer: a) using \ddpg  exploration with action perturbations to fill the buffer during 500 episodes (learning rate is zero); b) doing the same with \ddpg parameter perturbations; c) using samples collected from random policies (RP-PG). Appendix-\figurename~\ref{fig:sanity} compares these strategies to \ddpg with parameter perturbations and its corresponding \geppg version. Filling the replay buffer with exploration performed in the parameter space (b or c) seems not to impede \ddpg performance and even reduces variance. Finally, \geppg still outperforms all versions of \ddpg combined with undirected exploration strategies (2/2 tests positive on final metric, only bootstrap test on absolute metric).
\end{document}